\pdfoutput=1
\documentclass[11pt]{article}

\usepackage[final]{acl}

\usepackage{eso-pic} 
\RequirePackage{fancyhdr}
\RequirePackage{natbib}

\usepackage{times}
\usepackage{latexsym}

\usepackage[T1]{fontenc}
\usepackage[utf8]{inputenc}
\usepackage{inconsolata}
\usepackage{subfigure}
\usepackage{times}

\usepackage{xcolor}
\usepackage{soul}
\usepackage{latexsym}
\usepackage[T1]{fontenc}
\usepackage[utf8]{inputenc}
\usepackage{placeins}
\usepackage{inconsolata}
\usepackage{graphicx}
\usepackage{microtype}
\usepackage{hyperref}
\usepackage{url}
\usepackage{enumitem}
\usepackage{booktabs}
\usepackage{amsmath}
\usepackage{algorithm}
\usepackage{algpseudocode}
\usepackage{multirow}
\usepackage{multicol}
\usepackage{arydshln}
\usepackage{stfloats}
\usepackage{titlesec}
\usepackage{adjustbox}
\usepackage{changepage} %
\usepackage{lipsum} %
\usepackage{wrapfig}
\usepackage{lipsum}
\usepackage{listings}
\usepackage{tcolorbox}
\usepackage{amssymb} %
\usepackage{amsmath}
\usepackage{cuted}
\usepackage{caption}
\usepackage{subcaption}
\usepackage{siunitx}
\usepackage{array}
\usepackage{bbding}
\usepackage{fvextra}
\usepackage{bbm}
\usepackage{makecell} %
\usepackage[table]{xcolor} %

\definecolor{LightGreen}{rgb}{0.88,1,0.88}
\definecolor{LightRed}{rgb}{1,0.88,0.88}

\definecolor{sky}{RGB}{0, 230, 230}

\title{Graph-Reward-SQL: Execution-Free Reinforcement Learning for Text-to-SQL via Graph Matching and Stepwise Reward}

\author{
Han Weng\textsuperscript{1,2}\thanks{\fontsize{8.8pt}{10pt}\selectfont Work was done during the internship at ByteDance.},
Puzhen Wu\textsuperscript{2},
Longjie Cui\textsuperscript{2}\footnotemark[1],
Yi Zhan\textsuperscript{2}\footnotemark[1],
Boyi Liu\textsuperscript{2},\\
\textbf{Yuanfeng Song\textsuperscript{2}, Dun Zeng\textsuperscript{2}, 
Yingxiang Yang\textsuperscript{2},
Qianru Zhang\textsuperscript{2}},\\
\textbf{Dong Huang\textsuperscript{2,3}\footnotemark[2], %
Xiaoming Yin\textsuperscript{2},
Yang Sun\textsuperscript{2}\thanks{\fontsize{8.8pt}{10pt}\selectfont Corresponding authors
},
Xing Chen\textsuperscript{2}} \\
\textsuperscript{1} Beijing University of Posts and Telecommunications, China \\
\textsuperscript{2} ByteDance, China \\
\textsuperscript{3} Institute of Data Science, National University of Singapore \\
}

\begin{document}
\maketitle

\begin{abstract}

Reinforcement learning (RL) has been widely adopted to enhance the performance of large language models (LLMs) on Text-to-SQL tasks. However, existing methods often rely on execution-based or LLM-based Bradley–Terry reward models. The former suffers from high execution latency caused by repeated database calls, whereas the latter imposes substantial GPU memory overhead, both of which significantly hinder the efficiency and scalability of RL pipelines. To this end, we propose a novel reward model framework for RL-based Text-to-SQL named Graph-Reward-SQL, which employs the GMNScore outcome reward model. We leverage SQL graph representations to provide accurate reward signals while significantly reducing time cost and GPU memory usage. Building on this foundation, we further introduce StepRTM, a stepwise reward model that provides intermediate supervision over Common Table Expression (CTE) subqueries. This encourages both functional correctness and readability of SQL. Extensive comparative and ablation experiments on standard benchmarks, including Spider and BIRD, demonstrate that our method consistently outperforms existing reward models.
\end{abstract}

\section{Introduction}
Text-to-SQL \citep{tai2023exploring, li2024can, shi-etal-2025-gen} aims to translate natural language into structured database queries and plays a crucial role in democratizing data access by enabling non-technical users to interact with relational databases more effectively. A significant body of work has focused on fine-tuning foundational models, with recent studies showing that Reinforcement Learning (RL) can effectively enhance model performance \citep{pourreza2025reasoning, berdnyk2025llm, ma2025sql}. Among these efforts, the careful design of the reward model is a crucial challenge, as the quality of the reward signal directly influences policy optimization during fine-tuning.

In RL-based Text-to-SQL approaches, execution accuracy remains a dominant signal~\cite{nguyen2025fine, ma2025sql, pourreza2025reasoning, berdnyk2025llm}, providing intuitive feedback based on query correctness. 
Additionally, the Bradley–Terry reward model (BTRM) \cite{christiano2017deep} %
has been adapted for code generation by deriving preference pairs from execution outcomes \cite{AceCoder2025}. Structural rewards based on abstract syntax tree (AST) have also been explored to capture syntactic similarity~\cite{shojaee2023execution}. However, each approach has significant limitations in the Text-to-SQL tasks. Execution-based rewards introduce significant latency due to runtime database access. The LLM-based BTRM incurs high computational and GPU memory costs, which limits its scalability. AST matching-based similarity is prone to false negatives, where syntactically divergent queries that are semantically equivalent are penalized, leading to inaccurate reward signals. These limitations point to a key challenge in RL-based Text-to-SQL: designing an efficient reward model that can replace execution-based signals without compromising performance.

To address the above limitations, we introduce Graph-Reward-SQL, a novel reward model framework for RL-based Text-to-SQL. This framework incorporates two complementary reward models: Graph Matching Network Score (GMNScore) and Stepwise Relational Operator Tree Match (StepRTM). GMNScore serves as an outcome reward model, which evaluates the generated SQL queries using the Graph Matching Network (GMN) without requiring execution. GMN utilizes learned graph embeddings to assess functional equivalence, capturing the deep semantics of SQL queries~\cite{zhan-etal-2025-towards}. In contrast to execution-based rewards, GMNScore eliminates the need for costly database executions, resulting in a significant speed-up. Furthermore, compared to LLM-based Bradley-Terry reward models (BTRM), GMNScore substantially reduces GPU memory consumption due to the lightweight architecture of GMN. Additionally, StepRTM provides intermediate feedback through a stepwise reward mechanism that evaluates the generation of Common Table Expression (CTE) subqueries, complementing GMNScore.

The above design offers three notable advantages. (i) Superior Training Efficiency: Our method significantly reduces time cost and GPU memory usage compared to existing outcome reward models, leading to enhanced overall training efficiency for RL. (ii) Intermediate Feedback Integration: Unlike existing reward models that focus solely on outcome result, our framework incorporates intermediate evaluation by leveraging the structure of CTE SQL. This provides richer feedback during training, improving performance and readability. (iii) Strong Empirical Performance: Extensive ablation studies and evaluations on the Spider \cite{yu2018spider} and BIRD \cite{li2024can} Text-to-SQL benchmarks validate the superiority of our reward models. The results consistently demonstrate that our approach outperforms multiple strong reward model baselines, highlighting its effectiveness.

Our contributions are summarized as follows:
\begin{itemize}[leftmargin=*,itemsep=2pt,topsep=0pt,parsep=0pt]
    \item We propose GMNScore, an outcome reward model that leverages GMN to replace execution-based rewards, achieving both higher efficiency and better performance.
    \item We design a novel stepwise reward model StepRTM, which utilizes CTE SQL to deliver stepwise supervision by matching each subquery, resulting in improved accuracy and readability.
    \item Extensive experiments show that our reward models consistently improve performance while maintaining high inference efficiency and low GPU memory consumption.
\end{itemize}

\section{Related Work}
\noindent \textbf{Text-to-SQL.} Text-to-SQL is a key task in Natural Language Processing (NLP) that involves transforming queries expressed in natural language into executable SQL queries \citep{tai2023exploring, li2024can, shi-etal-2025-gen}. With the increasing deployment of large language models (LLMs), agentic frameworks \citep{wang-etal-2025-mac, pourreza2025chasesql, lei2024spider} have been introduced to enhance Text-to-SQL tasks. These frameworks enable LLMs to interact with databases through iterative reasoning and external tools. Code Foundation Models such as DeepSeek-Coder \citep{guo2024deepseek} and Qwen2.5-Coder \citep{Hui2024Qwen25CoderTR} provide the backbone for these agentic systems, enabling structured reasoning and code generation. Several approaches aim to improve LLM performance in Text-to-SQL tasks, including direct fine-tuning \citep{li2024codes, yang-etal-2024-synthesizing, pourreza-rafiei-2024-dts}, as well as techniques such as prompt design \citep{pourreza2023din, dong2023c3, promptgao2023}, self-consistency \citep{promptgao2023} and schema linking \citep{guo-etal-2019-towards, wang-etal-2020-rat, lei-etal-2020-examining, lee-etal-2025-mcs} to further optimize results.

\noindent \textbf{Reinforcement Learning and Reward Model.} 
RL has become an important paradigm for effectively fine-tuning Code Foundation Models. Policy optimization methods, such as Proximal Policy Optimization (PPO) \citep{schulman2017proximal} and Group Relative Policy Optimization (GRPO) \citep{shao2024deepseekmath}, have been explored. However, the effectiveness of RL training heavily relies on the quality of reward signals, making the design of reward models a critical aspect~\citep{trella2023reward}. 
Several contributions to RL-based code generation have advanced reward model strategies. Notable works include CodeRL \citep{le2022coderl}, which leverages execution feedback; PPOCoder \citep{shojaee2023execution}, integrating semantic matching of abstract syntax trees; and AceCoder \citep{AceCoder2025}, applying an LLM-based Bradley-Terry Reward Model.

The execution-based reward model for Text-to-SQL was initially used by \cite{Zhong2017Seq2SQLGS}. Recent advancements have introduced continuous reward scores based on keyword matching \citep{nguyen2025fine} and leveraged LLMs to generate reward function candidates and iteratively refine \citep{berdnyk2025llm}. Alongside these developments, reasoning models such as DeepSeek-R1 \citep{guo2025deepseek} have advanced RL in reasoning tasks, leading to the introduction of more sophisticated reward model designs. For example, SQL-R1 \citep{ma2025sql} incorporates format and length constraints, while Reasoning-SQL \citep{pourreza2025reasoning} employs more complex reward structures, such as schema linking feedback, n-gram similarity scores, and LLM-based judgment. Despite these enhancements, execution-based reward continue to play a central role in the above-mentioned approaches.

Current methods overlook the computational overhead of execution-based and LLM-based reward models and fail to fully exploit the deep semantic structure of SQL queries. Additionally, these approaches focus solely on evaluating the final generated SQL, neglecting the potential of leveraging intermediate supervision signals throughout the SQL generation process. To address these issues, we propose an execution-free outcome reward model and a stepwise reward mechanism. These methods significantly reduce computational overhead while providing more effective reward signals for RL-based Text-to-SQL tasks.

\section{Preliminaries}
\subsection{Problem Formulation}
In the standard Text-to-SQL setting, let $x$ denote a natural language query, $\hat{q}$ and $q^\star$ represent the generated SQL and reference SQL query, respectively. In this work, we mainly use Proximal Policy Optimization (PPO) \citep{schulman2017proximal}, which optimizes the policy model $\pi_{\theta}$ by maximizing:
$$
\begin{aligned}
\mathcal{J}(\theta) & = \mathbb{E}_{(x, \hat{q})\sim \mathcal{D}, \hat{q}\sim \pi_{\theta}(\cdot|x)} [ r(\hat{q}, q^\star) \\
& \quad - \beta \mathbb{D}_{\mathrm{KL}}(\pi_{\theta}(\cdot \mid x)\,\|\,\pi_{\mathrm{ref}}(\cdot \mid x))],
\end{aligned}
$$

\noindent where $\pi_{\mathrm{ref}}$ is the reference model, $\beta$ is a PPO hyperparameter and $r(\hat{q}, q^\star)$ is a reward model. Note that our method can be easily adapted to Group Relative Policy Optimization (GRPO) \cite{shao2024deepseekmath}, as detailed in the Appendix~\ref{appendix:GRPO}.

\subsection{Summary of Existing Reward Models}
Recognizing the great importance of reward models in RL, we discuss three types of main reward models. As summarized in Table~\ref{Table:RewardComparison}, we compare these models with our proposed reward models in terms of time cost and GPU memory usage during inference. Additionally, the final performance of all reward models is evaluated and ranked, as described in Section~\ref{exp:Reward Performance Comparison}. Detailed information on these comparisons can be found in the Appendix~\ref{appendix:cost}. 

\begin{table*}[ht]
\centering
\scalebox{0.8}{
\begin{tabular}{l|c|c|>{\centering\arraybackslash}p{2.5cm}|>{\centering\arraybackslash}p{2.5cm}|c}
\toprule
\rowcolor{gray!20}
\textbf{Reward Model} & \textbf{Modeling Basis} & \textbf{Granularity} & \textbf{Time Cost $\downarrow$} & \textbf{GPU Usage$\downarrow$} & \textbf{Perf. Rank (1=Best)}  \\
\midrule
EX \citep{pourreza2025reasoning} & Execution  & Outcome & \cellcolor{red!10}Slow & N/A & 3 \\
BTRM \citep{AceCoder2025} & LLM & Outcome & Moderate & \cellcolor{red!10}High & 5 \\
AstPM \citep{shojaee2023execution}  & AST Matching & Outcome & Fast & N/A & 6 \\
RelPM \citep{zhan-etal-2025-towards} & ROT Matching & Outcome & Fast & N/A & 4 \\
\cmidrule{1-6}
GMNScore (Ours) & GMN & Outcome & \cellcolor{green!10}Fast & \cellcolor{green!10}Low & 2 \\
+ StepRTM (Ours) & ROT Matching & Stepwise & Fast & N/A & \textbf{1}\\
\bottomrule
\end{tabular}
}
\vspace{-5pt}
\caption{
Comparison of Reward Models in RL for Text-to-SQL Tasks. Our proposed GMNScore and StepRTM achieve better performance while significantly reducing time and memory costs.}
\vspace{-10pt}
\label{Table:RewardComparison}
\end{table*}

\paragraph{Execution Accuracy (EX).}
For the Text-to-SQL tasks, the execution accuracy serves as the most direct reward signal, providing a discrete score based on whether the generated SQL query yields the correct result upon execution. We use a discrete reward model with finer-grained feedback based on syntax error \cite{pourreza2025reasoning} and runtime diagnostics following \cite{shojaee2023execution}. Given a generated SQL $\hat{q}$ and reference SQL $q^\star$, the formulation is listed as: 

\vskip-4mm 
$$r_{\text{EX}}(\hat{q},q^\star) =R_{exec} + R_{syntax} + R_{runtime}$$

However, the EX has notable limitations. When the database contains poor quality data (e.g., limited, missing, or inconsistent entries) or structural issues (e.g., redundancy or anomalies), different queries may produce identical results \citep{zhong2020semantic}. Test Suite (TS) \citep{zhong2020semantic} attempted to address this issue, but as shown in \cite{zhan-etal-2025-towards}, false positives and false negatives remain unavoidable. Additionally, repeatedly executing SQL queries introduces significant computational overhead, increasing training time. More details about EX are provided in the Appendix~\ref{appendix:ex}.

\paragraph{Bradley-Terry Reward Model (BTRM).}
Given a natural language input $x$ and a candidate SQL query $y$, we define the reward model as $r_\psi(x,y) = h_r\bigl(\mathcal{M}_\theta(x,y)\bigr),$
with a pretrained language model $\mathcal{M}_\theta$ and a reward head $h_r$. The training process uses preference pairs based on execution correctness: $\mathcal{D} = \{(x_i, y_i^+, y_i^-)\}_{i=1}^N,$ where $ y_i^+$ executes correctly and $y_i^-$ fails or returns an incorrect result \citep{zeng2025acecoder}. The objective is to minimize the Bradley-Terry log-likelihood \cite{bradley1952rank} as follows:
$$
\begin{aligned}
-\sum_{i=1}^N
\log&
\frac{\exp\bigl(r_\psi(x_i,y_i^+)\bigr)}
 {\exp\bigl(r_\psi(x_i,y_i^+)\bigr) + \exp\bigl(r_\psi(x_i,y_i^-)\bigr)}
 \end{aligned}
$$
This model learns to assign higher scores to correct queries, providing a dense proxy reward model for RL \cite{christiano2017deep}. In contrast to EX, BTRM enables more efficient policy training by eliminating the need to query databases. However, the large parameter size of LLM-based BTRM significantly increases GPU memory usage.

\paragraph{Matching-based Reward.}
In \citep{nguyen2025fine}, rule-based keyword matching is used for scoring SQL queries, while n-gram similarity is used in Reasoning-SQL \citep{pourreza2025reasoning} to capture overlapping token sequences. Matching-based methods are fast and model-free, but may assign negative rewards to semantically equivalent SQL queries that differ in syntax, which should ideally be considered correct. In broader code generation tasks, the PPOCoder \citep{shojaee2023execution} uses semantic matching of abstract syntax trees and data flow graphs. However, it still focuses on surface-level structure and does not fully capture the deep semantic information.

\section{Methodology}
We introduce Graph-Reward-SQL, a novel reward model framework designed to enhance SQL generation through two key innovations. First, we propose GMNScore, which replaces EX, reducing time costs while maintaining the accuracy of reward signals without requiring database execution. Second, we introduce StepRTM, a stepwise reward model based on the Relational Operator Tree (ROT) representation of CTE SQL, which provides supplementary intermediate feedback.

\begin{figure*}[t]
    \centering
    \includegraphics[width=1\linewidth]{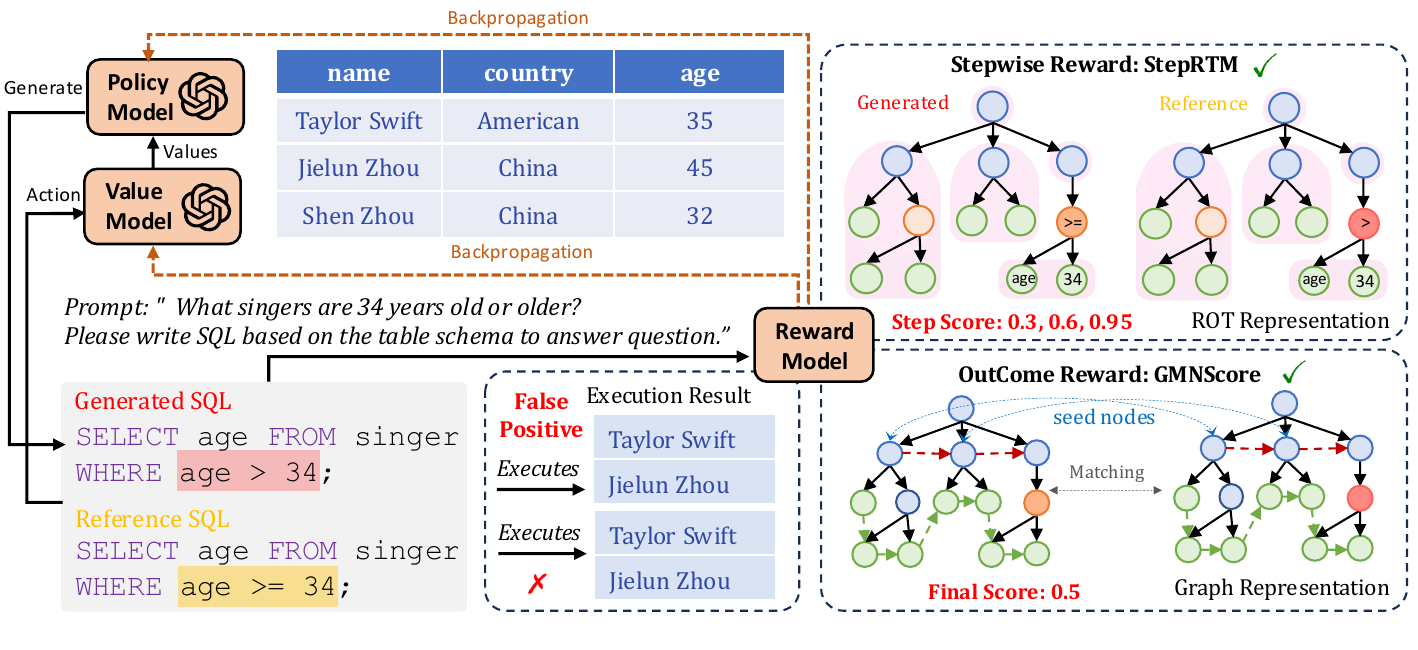}
    \vspace{-25pt}
    \caption{Graph-Reward-SQL employs PPO, where rewards drive policy updates. An example illustrates a limitation of EX: the generated SQL 'WHERE age > 34' and the reference SQL 'WHERE age >= 34' produce identical results despite their semantic difference. In contrast, our proposed GMNScore leverages graph representation to capture deep semantic similarity. Additionally, a stepwise reward model, StepRTM, that is tailored for CTE SQL addresses the lack of intermediate rewards. A mock stepwise score is provided for illustration; see Figure~\ref{fig:step_relpm_pipeline} for more details.
}\label{fig:pipeline}
\vspace{-8pt}
\end{figure*}

\subsection{Relational Operator Tree (ROT)}
Accurately modeling SQL structure and semantics is crucial for query analysis and comparison. SQL queries can be converted into Abstract Syntax Trees (ASTs) to capture their syntactic structure. However, unlike general programming languages, SQL lacks key representations like Control Flow Graphs (CFGs) \citep{cota1994control} and Data Flow Graphs (DFGs) \citep{orailoglu1986dataflowgraphrepresentation}, which are essential for reflecting logic and data dependencies.

To bridge this gap, we leverage the Relational Operator Tree (ROT) to represent SQL queries as trees of relational algebra operators. Each node in the tree corresponds to a specific logical operation (e.g., \texttt{Join}, \texttt{Project}, \texttt{Filter}), while the tree structure itself reflects the dependencies and execution order of the query. In practice, we use Apache Calcite \citep{begoli2018apache} to generate ROTs, which compiles SQL into a canonical intermediate representation called \texttt{RelNode}. This format includes various optimizations, such as operator reordering and clause simplification, resulting in normalized logical plans that are more resilient to surface-level differences. Similar to CFGs and DFGs, the RelNode format can also integrate control dependencies and data flow as edges \citep{zhan-etal-2025-towards}. This enables the creation of more comprehensive graph representations that include richer SQL semantics, facilitating query understanding and evaluation.

\subsection{FuncEvalGMN}
\label{funceval_gmn}
After obtaining the SQL graph representations $G_1$ and $G_2$, we employ a Graph Matching Network (GMN) \citep{li2019graph} trained in SQL pairs \citep{zhan-etal-2025-towards} to assess functional equivalence. It is trained using contrastive learning for pretraining and supervised learning to capture deep semantic similarity of SQL queries. The similarity between two queries is computed as the negative Euclidean distance between their final graph-level embeddings: $s(h_{G_1}, h_{G_2}) = \big\lVert h_{G_1} - h_{G_2} \big\rVert_2,$ where $h_{G_1}$ and $h_{G_2}$ are computed by the GMN, considering the joint representations of $G_1$ and $G_2$. This approach, first introduced in FuncEvalGMN \citep{zhan-etal-2025-towards}, is described in further detail in Appendix~\ref{appendix:funcevalgmn_detail}, and the details of our further optimization are provided in Appendix~\ref{appendix:gmn_train}.

\subsection{ROT/RelNode Partial Matching (RelPM)}
Similar to AST, RelNode can also be used to evaluate SQL similarity through graph matching. RelPM \citep{zhan-etal-2025-towards} is a rule-based matching algorithm that assesses the similarity of SQLs based on their RelNode representations, denoted $\mathcal{G}_{\hat{q}}$ and $\mathcal{G}_{q^\star}$, respectively. A comparable approach, applied to AST structures, is known as AstPM \citep{zhan-etal-2025-towards}. Both algorithms adopt a hierarchical partial matching strategy and derive a global similarity score based on the Precision and Recall of node-level matching results. At the node level, matches are determined by comparing each generated node $n'\in \mathcal{G}_{\hat{q}}$ with all the candidate nodes $n \in \mathcal{G}_{q^\star}$ in the reference tree. A match is established when two nodes have the same operator type and value. Additionally, a matching score is computed by comparing their subgraphs, and the candidate node with the highest matching score is selected as the final match. Further details are provided in Appendix~\ref{appendix:relpm_detail}.

\begin{figure*}[ht]
    \centering
\includegraphics[width=1\linewidth]{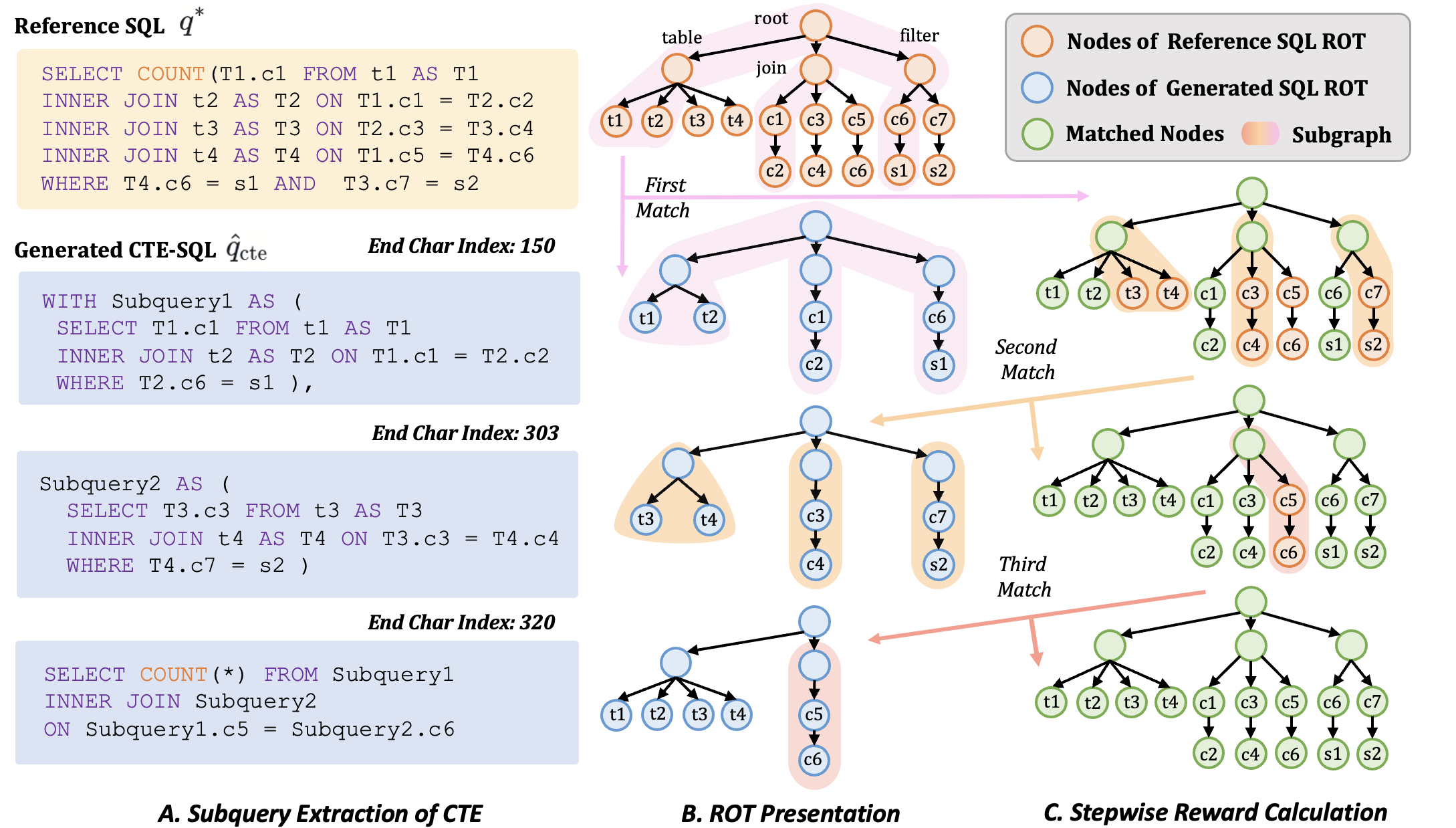}
    \vspace{-15pt}
    \caption{
Overview of the StepRTM Stepwise Reward Calculation. (a) The generated SQL $\hat{q}_{\text{cte}}$ is segmented into a sequence of subqueries, with the end index of each subquery recorded. (b) Both the reference SQL query $q^*$ and each subquery are parsed into ROTs (c) A stepwise matching process is performed between the ROTs. At each step, newly matched nodes are identified and used to compute incremental rewards.}
    \label{fig:step_relpm_pipeline}
\vspace{-10pt}
\end{figure*}

\subsection{Reward Function Design}\label{section:reward_design}
Figure~\ref{fig:pipeline} illustrates our reward design, comprising the outcome reward model GMNScore and the stepwise model StepRTM. Given the generated SQL $\hat{q}$ and the reference SQL $q^\star$, the reward at time-step $t$ for a sequence of length $T$ is computed as follows:
\vskip-5mm
$$
\label{eq:comreward}
\resizebox{1\linewidth}{!}{$
\begin{aligned}
\mathcal{R}&_t(\hat{q},q^*) = \;\mathbbm{1}(\textit{cond}_{\text{eos}}) \cdot \left[ R_{\text{GMNScore}}(\hat{q}, q^*) - \beta R_{\text{kl}}(\hat{q}_{<t}) \right] \\
& + \mathbbm{1}(\textit{cond}_{\text{sub}}) \cdot \left[ R_{\text{StepRTM}}(\hat{q}_{\leq t}, q^*) - \beta R_{\text{kl}}(\hat{q}_{<t}) \right] \\
& + \mathbbm{1}(\neg \textit{cond}_{\text{eos}}) \cdot \mathbbm{1}(\neg \textit{cond}_{\text{sub}}) \cdot \left[ - \beta R_{\text{kl}}(\hat{q}_{<t}) \right],
\end{aligned}
$}
$$
\noindent where $\textit{cond}_{\text{eos}}$ indicates the end of generation, at which point the outcome reward model $R_{\text{GMNScore}}$ is applied. $\textit{cond}_{\text{sub}}$ signifies the completion of a subquery, triggering the stepwise reward model $R_{\text{StepRTM}}$ to compare the current subquery with the corresponding substructure in the reference query. The symbol $\neg$ denotes logical negation. $R_{\text{kl}}(\hat{q}_{<t})$  represents a KL-divergence penalty that measures the deviation between the learned policy and the pretrained language model, applied at each time step to regularize policy updates. The scalar $\beta$ is a hyperparameter that balances rewards with policy regularization.

\subsection{Outcome Reward: GMNScore}
As described in Section~\ref{funceval_gmn}, the functional correctness of generated SQL can be evaluated using the FuncEvalGMN metric $\mathcal{M}_{\text{GMN}}$, which aligns well with the objective of reward model in RL. We design an outcome reward model as follows: 
\vskip-4mm 
$$
\resizebox{1\linewidth}{!}{$
\begin{aligned}
R_\text{GMNScore} (\hat{q},q^\star) =
\begin{cases}
-1, & \text{if syntax error}\\
-0.6, & \text{if ROT error}\\
\max(0, \mathcal{M}_{\text{GMN}} + 1) &
\end{cases}
\end{aligned}
$}
$$
The GMNScore formulation introduces graded penalties for SQL queries that trigger syntax errors or ROT parsing errors\footnote{Refers to failures in converting SQL into a ROT using Apache Calcite. For example, the query \texttt{WITH sub1 AS (SELECT id, name FROM users) SELECT age FROM sub1;} leads to an error “Column `age` not found in any table”.}. For all other cases, we rescale the similarity metric $\mathcal{M}_{\text{GMN}}$ (which lies in the range $(-\infty, 0]$) to the interval $[0, 1)$ by first applying an affine shift and then rectifying any negative values to zero.

\subsection{Stepwise Reward: StepRTM}
Current ETL (Extract, Transform, Load) pipelines rarely execute their logic in a single step. Instead, analysts break the workflow into a detailed plan of subqueries, where each subquery progressively transforms the data until the query is complete. This logic is typically expressed using CTEs, as demonstrated by the simplified example below:

\begin{lstlisting}[language=SQL,basicstyle=\ttfamily\small,  keywordstyle=\bfseries,morekeywords={WITH}]
WITH step1 AS (/* subquery1 */), 
     step2 AS (/* subquery2 */)
SELECT ... FROM step2;
\end{lstlisting} 
In most cases, CTEs enhance the readability of complex SQL by providing clear representations of intermediate steps in an ETL pipeline. These steps not only facilitate data transformation but also offer a natural way to evaluate the process stepwise.

Inspired by subgraph matching techniques \citep{lou2020neural, Roy_Velugoti_Chakrabarti_De_2022}, we propose Stepwise Relational Operator Tree Matching (StepRTM), which incorporates stepwise reward scores to provide intermediate feedback. The overall procedure of StepRTM is illustrated in Figure~\ref{fig:step_relpm_pipeline}. Let $q^*$ denote the reference SQL, and represent the generated SQL as a sequence of subqueries $\hat{q}_{\text{cte}} = [\hat{q}_1, \hat{q}_2, \ldots, \hat{q}_n]$. Let $\mathcal{G}_{q^*}$ and $\mathcal{G}_{\hat{q}_i}$ denote the node sets of the ROT representations for the reference query and the $i$-th generated subquery. The stepwise scores are then computed as follows: 

\vskip-4mm 
$$
\begin{aligned}
&\mathcal{R}_{\text{StepRTM}}^{(i)}(\hat{q}_{\text{cte}}, q^*)=\frac{\bigl|\bigl(\mathcal{M}_i \cup \mathcal{G}_i\bigr)\,\cap\,\mathcal{G}_{q^*}\bigr|}{|\mathcal{G}_{q^*}|},
\end{aligned}
$$

\noindent where $\mathcal{M}_{i} = \bigcup_{j=1}^{i-1} \mathcal{G}_j$ represents all the matched subgraphs parsed from the first $i$ subqueries, $\mathcal{G}_j$ denotes the maximal matched subgraph in the reference query that aligns with the $i$-th subquery $\hat{q}_i$. This formulation prevents repeated rewards for the same reference node and ensures that the overall signal reflects the incremental semantic coverage of the target reference query. Stepwise supervision improves training performance by providing richer intermediate feedback, facilitating the generation of correct SQL queries.

\section{Experimental Setup}\label{sec::exp}

\paragraph{Datasets.}\label{sec::dataset}
Our experiments are primarily conducted on the Spider and BIRD benchmarks. The Spider dataset \citep{yu2018spider} contains 10,181 natural language questions paired with 5,693 complex SQL queries across 138 domains. The BIRD dataset \citep{li2024can} consists of 12,751 questions spanning more than 37 professional fields. We used the training split of the Spider dataset for training and the development splits of Spider and BIRD for evaluation. Additionally, we used a subset of the 200k-Text2SQL dataset\footnote{\url{https://huggingface.co/datasets/philikai/200k-Text2SQL}} in a warm-up phase prior to RL training. Further details about the datasets are provided in Appendix~\ref{appendix:datasets}.

\paragraph{Baselines.}
We benchmark against representative state-of-the-art (SOTA) Text-to-SQL pipelines spanning in-context learning, self-correction, post-training, and multi-agent paradigms \citep{
  pourreza2023din,           %
  promptgao2023,             %
  pourreza-rafiei-2024-dts,  %
  gao2024xiyan,              %
  pourreza2025chasesql,      %
  pourreza2025reasoning,     %
  Zhang2025RewardSQLBT}. 
We further compare three widely adopted reward model designs in RL training to validate the effectiveness of the proposed reward models: (i) the execution-based reward EX, widely used in recent studies \citep{nguyen2025fine, berdnyk2025llm, ma2025sql, pourreza2025reasoning}; (ii) an LLM-based Bradley–Terry reward model (BTRM) \citep{christiano2017deep, zeng2025acecoder}, trained using the DeepSeek-Coder-1.3B-Ins as the backbone, as detailed in Appendix~\ref{appendix:RLHF_LLM_training}, to evaluate the efficacy of model-based reward mechanisms; and (iii) AstPM and RelPM \citep{zhan-etal-2025-towards}, which, motivated by recent works \citep{shojaee2023execution, nguyen2025fine, pourreza2025reasoning}, are incorporated as matching-based reward model baselines.

\paragraph{Evaluation Metrics.} 
\label{sec::mertic}
Consistent with prior work \citep{pourreza2025chasesql,pourreza2025reasoning,Zhang2025RewardSQLBT}, we adopt Execution Accuracy (EX) as the primary metric for comparisons with state-of-the-art systems. In addition, we report Test-Suite Accuracy (TS), which evaluates predicted queries against suites of distilled or fuzzed database variants under varied conditions \citep{zhong2020semantic}. Since TS mitigates the false positives that may arise from insufficiently discriminative databases, it provides a more robust estimate of semantic correctness \citep{promptgao2023,li2024codes,yang-etal-2024-synthesizing}. We therefore use TS to more accurately quantify the gains achieved by our reward model.

\paragraph{Implementation Details.} 
All experiments are conducted on several edge servers. We extend the verl\footnote{\url{https://github.com/volcengine/verl}} and DeepSpeed-Chat\footnote{\url{https://github.com/deepspeedai/DeepSpeed}} \citep{yao2023dschat} framework to support a comparison of multiple reward models, including execution-based, LLM-based, and matching-based rewards.

Prior to RL training, we performed SFT using two cold-start datasets. First, we sampled a subset from the 200k-Text2SQL dataset, matching the size of the Spider training set, and trained the base model for two epochs. To promote the generation of CTE SQL queries in the stepwise reward PPO experiments, we converted BIRD data into CTE format to prepare a warm-up dataset referred to as CTE-SFT. During supervised fine-tuning (SFT), we employ the AdamW optimizer (initial learning rate 5e-6, batch size 4) and use a polynomial scheduler to decay the learning rate throughout training. Additional details
about hyperparameter of SFT and RL are provided in Appendix~\ref{appendix:hyperparameter_setting}.

\section{Results}\label{sec::result}

\begin{table*}[ht!]
\centering
\resizebox{\textwidth}{!}{
\begin{tabular}{l p{5.6cm} l c c c}
\toprule
\textbf{Method} & \textbf{Method Detail} & \textbf{Base Model} & \textbf{Spider EX} & \textbf{BIRD EX} & \textbf{Avg.} \\
\midrule
DIN-SQL \citep{pourreza2023din} & In-context learning & GPT-4 & 82.88 & 50.72 & 66.80 \\
DAIL-SQL \citep{promptgao2023}  & In-context learning & GPT-4 & 74.47 & 54.76 & 64.62 \\
DTS-SQL \citep{pourreza-rafiei-2024-dts} & Schema-linking, SFT & DeepSeek-7B & - & 55.80 & - \\
XiYan-SQL \citep{gao2024xiyan} & Multi-Agent & QwenCoder-32B & - & 67.01 & - \\
CHASE-SQL \citep{pourreza2025chasesql} & Multi-Agent & Gemini-1.5-pro & - & 74.46 & - \\
\midrule
Reasoning-SQL \citep{pourreza2025reasoning}  & Multi-Agent, GRPO & Qwen2.5-Coder-7B & 78.72 & \textbf{64.01} & 71.37 \\
Reward-SQL \citep{Zhang2025RewardSQLBT} & GRPO & Qwen2.5-Coder-7B & 77.08 & 59.70 & 68.39 \\
\textbf{Graph-Reward-SQL (Ours)} & GRPO & Qwen2.5-Coder-7B & \textbf{81.62} & 63.04 & \textbf{72.33} \\
\bottomrule
\end{tabular}
}

\caption{Performance comparison of Graph-Reward-SQL and baseline models on the Spider and BIRD dataset. Baseline results are reported from their original publications.}
\vspace{-10pt}
\label{tab:sotas}
\end{table*}

We benchmark our method against recent SOTA Text-to-SQL systems, including in-context learning and multi-agent approaches. As shown in Table~\ref{tab:sotas}, Graph-Reward-SQL reaches 81.62\% EX on Spider and 63.04\% on BIRD trained with GRPO on the Qwen2.5-Coder-7B-Ins model. Our reward model framework delivers performance competitive with the Reasoning-SQL, even as the latter employs a multi-agent system and five reward models that incorporate both EX and LLM-based judgment.

\subsection{Reward Performance Comparison}
\label{exp:Reward Performance Comparison}

\begin{table}[t!]
\centering
\scriptsize
\renewcommand{\arraystretch}{1.2}
\resizebox{0.5\textwidth}{!}{
\begin{tabular}{llccc}
\toprule
\textbf{Method} & \textbf{Spider} & \textbf{BIRD} & \textbf{Avg.} \\
\midrule
\textbf{DeepSeek-Coder-1.3B-Ins} & 39.56 & 11.34 & 25.45 \\
 + SFT & 57.74 & 13.30 & 35.52 \\
 + PPO w/ AstPM & 59.96 & 12.52 & 36.24 \\
 + PPO w/ RelPM & 62.86 & 14.67 & 38.77 \\
 + PPO w/ BTRM & 61.41 & 14.21 & 37.81 \\
 + PPO w/ EX & 65.28 & \textbf{17.21} & 41.25 \\
 + PPO w/ GMNScore (Ours) & \textbf{67.70} & 16.10 & \textbf{41.90} \\
\midrule
\textbf{DeepSeek-Coder-6.7B-Ins}  & 44.97 & 18.38 & 31.68 \\
 + SFT & 69.05 & 21.71 & 45.38 \\
 + PPO w/ AstPM & 70.31 & 22.88 & 46.60 \\
 + PPO w/ RelPM & 70.60 & 26.01 & 48.31 \\
 + PPO w/ BTRM & 67.99 & 22.23 & 45.11 \\
 + PPO w/ EX & 71.66 & \textbf{26.66} & 49.16 \\
 + PPO w/ GMNScore (Ours) & \textbf{72.44} & 26.14 & \textbf{49.29} \\
\bottomrule
\end{tabular}
}
\caption{TS Performance of Deepseek-Coder-1.3B-Ins and Deepseek-Coder-6.7B-Ins models under multiple baselines and proposed GMNScore outcome reward.}
\label{tab:results}
\end{table}

\textbf{GMNScore can replace the EX, thereby eliminating dependence on SQL execution and database environments.} As demonstrated in Table \ref{tab:results}, GMNScore achieves the highest average TS for the 1.3B and 6.7B models, highlighting the importance of well-designed reward signals in RL. Another notable observation is that RelPM outperforms AstPM, with improvements of 2.53\% and 1.71\% for the two model sizes, respectively. The better performance of the former can be attributed to the use of normalized logical plans for SQL parsing in ROT, which are less susceptible to surface-level syntactic differences. ROT also provides an effective representation way for our proposed GMNScore and StepRTM reward models. GMNScore learns deep semantic information via graph-level embeddings, bypassing the need for execution-result comparisons and thus mitigating false-positive noise. Additionally, GMNScore eliminates the necessity of constructing and maintaining databases, offering a lightweight solution for large-scale RL-based Text-to-SQL. Case studies are provided in Appendix ~\ref{appendix:case_study}.

\begin{table}[t!]
\centering
\resizebox{1\linewidth}{!}{
\begin{tabular}{lccc}
\toprule
\textbf{Method} & \textbf{Spider} & \textbf{BIRD} & \textbf{Avg.} \\
\midrule
\textbf{DeepSeek-Coder-1.3B-Ins + PPO} & 39.56 & 11.34 & 25.45 \\
RelPM w/ SFT & 62.86 & 14.67 & 38.77 \\
RelPM w/ CTE-SFT & 62.57 & 19.36 & 40.97 \\
RelPM \& StepRTM w/ CTE-SFT & \textbf{64.31} & \textbf{19.43} & \textbf{41.87} \\
\midrule
EX w/ SFT & 65.28 & 17.21 & 41.25 \\
EX w/ CTE-SFT & 65.09 & 18.97 & 42.03 \\
EX \& StepRTM w/ CTE-SFT & \textbf{67.89} & \textbf{19.75} & \textbf{43.82} \\
\midrule
GMNScore w/ SFT & 67.70 & 16.10 & 41.90 \\
GMNScore w/ CTE-SFT & 68.57 & 20.40 & 44.49 \\
GMNScore \& StepRTM w/ CTE-SFT & \textbf{68.67} & \textbf{21.97} & \textbf{45.32} \\
\bottomrule
\end{tabular}
}
\caption{TS performance of the DeepSeek-Coder-1.3B-Ins model trained with StepRTM integration. Both SFT and CTE-SFT refer to different warm-up datasets.}
\vspace{-10pt}
\label{tab:StepRTM-exp}
\end{table}

\textbf{The integration of StepRTM as a stepwise reward further enhances performance.} As shown in Table \ref{tab:StepRTM-exp}, combining CTE-SFT with StepRTM consistently results in metric improvements across various outcome reward models. Notably, our framework, which integrates GMNScore alongside StepRTM, achieves the highest result. Specifically, we observe a 5.87\% improvement on the BIRD dataset and a 0.97\% increase on the Spider dataset. These observations suggest that the BIRD dataset, which is inherently more challenging due to its diverse database and query complexity, benefits more significantly from our proposed stepwise reward.

\subsection{Effectiveness of GMNScore with GRPO}

\begin{figure}[t]
\centering
\includegraphics[width=1\linewidth]{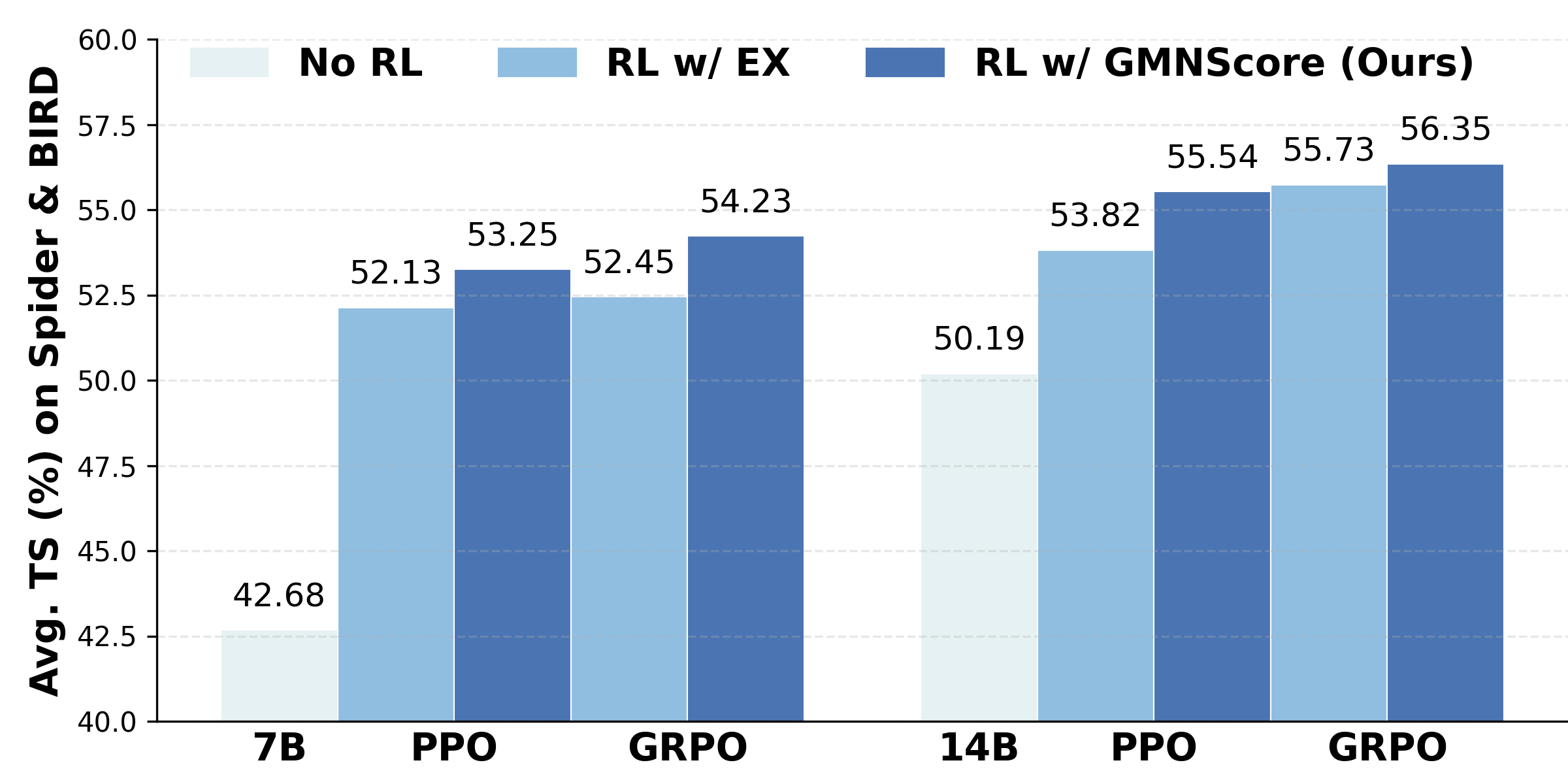}
\vspace{-10pt}
\caption{TS Performance of Qwen2.5-Coder-7B/14B-Ins models directly trained by PPO/GRPO.}
\label{fig:qwen_performance}
\end{figure}

\textbf{GMNScore performs robustly across both PPO and GRPO protocols.} As shown in Figure~\ref{fig:qwen_performance}, the results consistently demonstrate that GMNScore outperforms EX in these two RL protocols, underscoring its robustness and effectiveness. We report the average performance across two benchmarks, with detailed results provided in Appendix~\ref{appendix:exp_grpo_result}.

\begin{table}[ht]
\centering
\small
\begin{tabular}{lccc}
\toprule
\textbf{Reward} & \textbf{Time$\downarrow$} & \textbf{Params$\downarrow$} & \textbf{GPU Memory$\downarrow$} \\
\midrule
EX & 1.088s & -- & -- \\
BTRM & 0.095s & 1.35B & 9304MB \\
GMNScore & \textbf{0.023s} & \textbf{3.99M} & \textbf{83MB} \\
\bottomrule
\end{tabular}
\caption{Average per-sample cost of reward scoring.}
\label{tab:reward_cost_body}
\vspace{-10pt}
\end{table}

\subsection{Cost of GMNScore}
We compare the cost of different reward models by running one PPO epoch and measuring the average reward-scoring latency per sample. As shown in Table~\ref{tab:reward_cost_body}, GMNScore is the most efficient. In contrast, EX is the slowest due to repeated database calls. From a model-size perspective, GMNScore is highly compact, with 3.99M parameters, whereas the widely used BTRM is substantially larger at 1.35B parameters. GMNScore is trained once offline and can be reused across multiple RL experiments, which significantly reduces overall computational overhead. Additional details on the measurement setup are provided in Appendix~\ref{appendix:cost}.

\begin{table*}[t!]
\centering
\scriptsize
\renewcommand{\arraystretch}{1}
\resizebox{\textwidth}{!}{
\begin{tabular}{p{0.30\linewidth} p{0.30\linewidth}p{0.5\linewidth}}
\toprule
\textbf{Reference SQL} & \textbf{Failed SQL \textcolor{red}{\XSolidBrush}} & \textbf{CTE SQL \textcolor{green!50!black}{\checkmark}} \\
\midrule
\multicolumn{3}{l}{\textbf{Question: What is the average score of Stephen Turner's posts?}} \\
\midrule
\parbox[t]{\linewidth}{\ttfamily
SELECT AVG(T2.Score) \\
FROM \sethlcolor{LightGreen}\hl{users} AS T1 \\
INNER JOIN \sethlcolor{LightGreen}\hl{posts} AS T2 ON T1.Id = T2.OwnerUserId \\
WHERE T1.DisplayName = \sethlcolor{LightGreen}\hl{'Stephen Turner'}
}
& \parbox[t]{\linewidth}{\ttfamily
SELECT AVG(T1.Score) \\
FROM \sethlcolor{LightRed}\hl{comments} AS T1 \\
INNER JOIN \sethlcolor{LightGreen}\hl{users} AS T2 ON T1.UserID = T2.ID \\
WHERE T2.DisplayName = \sethlcolor{LightGreen}\hl{'Stephen Turner'}
}
& \parbox[t]{\linewidth}{\ttfamily
WITH UserInfo AS ( SELECT id FROM \sethlcolor{LightGreen}\hl{users} \\
\quad WHERE displayname = \sethlcolor{LightGreen}\hl{'Stephen Turner'}), \\
PostInfo AS ( SELECT score FROM \sethlcolor{LightGreen}\hl{posts} \\
\quad WHERE owneruserid IN (SELECT id FROM UserInfo)) \\
SELECT AVG(score) FROM PostInfo
}
\\ \midrule

\multicolumn{3}{l}{\textbf{Question: List down at least five superpowers of male superheroes.}} \\
\midrule
\parbox[t]{\linewidth}{\ttfamily
SELECT T3.power\_name \\
FROM \sethlcolor{LightGreen}\hl{superhero} AS T1 \\
INNER JOIN \sethlcolor{LightGreen}\hl{hero\_power} AS T2 ON T1.id = T2.hero\_id \\
INNER JOIN \sethlcolor{LightGreen}\hl{superpower} AS T3 ON T3.id = T2.power\_id \\
INNER JOIN \sethlcolor{LightGreen}\hl{gender} AS T4 ON T4.id = T1.gender\_id \\
WHERE T4.\sethlcolor{LightGreen}\hl{gender} = 'Male' \\
\sethlcolor{LightGreen}\hl{LIMIT 5}
}
& \parbox[t]{\linewidth}{\ttfamily
SELECT T1.power\_name \\
FROM \sethlcolor{LightGreen}\hl{superpower} AS T1 \\
INNER JOIN \sethlcolor{LightGreen}\hl{hero\_power} AS T2 ON T1.id = T2.power\_id \\
INNER JOIN \sethlcolor{LightGreen}\hl{superhero} AS T3 ON T3.id = T2.hero\_id \\
WHERE T3.\sethlcolor{LightRed}\hl{gender\_id} = 1 \\
\sethlcolor{LightRed}\hl{GROUP BY T1.power\_name} \\
\sethlcolor{LightGreen}\hl{LIMIT 5}
}
& \parbox[t]{\linewidth}{\ttfamily
WITH MaleSuperheroes AS ( SELECT id \\
\quad FROM \sethlcolor{LightGreen}\hl{superhero} \\
\quad WHERE gender\_id IN ( \\
\quad\quad SELECT id FROM \sethlcolor{LightGreen}\hl{gender} \\
\quad\quad WHERE \sethlcolor{LightGreen}\hl{gender} = 'Male'), \\
SuperpowersOfMaleSuperheroes AS ( SELECT DISTINCT T1.power\_name \\
\quad FROM \sethlcolor{LightGreen}\hl{superpower} AS T1 \\
\quad INNER JOIN \sethlcolor{LightGreen}\hl{hero\_power} AS T2 ON T1.id = T2.power\_id \\
\quad INNER JOIN MaleSuperheroes AS T3 ON T3.id = T2.hero\_id) \\
SELECT power\_name FROM SuperpowersOfMaleSuperheroes  \sethlcolor{LightGreen}\hl{LIMIT 5}
}
\\
\bottomrule
\end{tabular}
}
\caption{Comparisons among Reference SQL, Failed SQL, and CTE SQL demonstrate the effectiveness of StepRTM.}
\vspace{-10pt}
\label{tab:case_comparison}
\end{table*}

\subsection{Intrinsic Evaluation of Reward Models}

\begin{table}[t!]
\centering
\resizebox{1\linewidth}{!}{
\begin{tabular}{lcc}
\toprule
\textbf{Reward Model} & \textbf{Spider-pair dev} & \textbf{BIRD-pair dev} \\
\midrule
AstPM & 82.81\% & 80.38\% \\
RelPM & 84.42\%  & 83.57\%  \\
BTRM & 89.24\%  & 84.77\%\\
EX & 96.37\%  & 92.67\%  \\
GMNScore & \textbf{97.62\%} & \textbf{94.14\%} \\
\bottomrule
\end{tabular}
}
\caption{Area Under the Curve (AUC) of reward models on Spider-dev-pair and BIRD-dev-pair.}
\label{tab:reward_auc}
\vspace{-10pt}
\end{table}

Area Under the Curve (AUC) is used as a statistically principled measure of similarity-evaluation accuracy \citep{zhan-etal-2025-towards}; its alignment with RL reward-model objectives makes it a rigorous criterion for validating reward-model performance. As shown in Table~\ref{tab:reward_auc}, GMNScore achieves 97.62\% AUC on Spider-dev-pair and 94.14\% on BIRD-dev-pair. Equivalence labels in the these two dev-pair sets were manually verified to minimize false positives, improving the reliability of these AUC estimates. Collectively, the results indicate that GMNScore provides accurate and consistent preference judgments, laying a solid foundation for subsequent RL training. Details on training and evaluation datasets are provided in Appendix~\ref{appendix:gmn_train}.

\subsection{Case Study: CTE SQL with StepRTM}  
\label{cte_casestudy} 
Table~\ref{tab:case_comparison} presents two cases that demonstrate how the stepwise reward model enhances both correctness and readability. Each case compares the reference SQL, a failed SQL query generated by a model trained solely with an outcome reward model, and a CTE SQL query generated by a model trained with StepRTM. In the first case, the failed SQL incorrectly retrieves data from the ‘\texttt{comments}’ table instead of the intended ‘\texttt{posts}’ table. The CTE SQL resolves this by decomposing the task into clear subqueries: first locating the target user, then aggregating the scores of that user's posts. In the second case, the failed SQL hard-codes the gender identifier as '\texttt{1}', leading to errors in filtering. In contrast, the CTE SQL uses two dedicated subqueries to correctly filter male superheroes, extract their superpowers, finally combine the results.

\section{Discussion}
The GMNScore introduced in this paper offers an alternative to EX while remaining fully compatible with other reward models. As detailed in Appendix~\ref{appendix:hybrid_outcome}, we extend our investigation beyond the StepRTM integration (Section~\ref{exp:Reward Performance Comparison}) by applying hybrid outcome reward models, which further improve performance. This finding is consistent with previous work using multiple outcome rewards \citep{pourreza2025reasoning, ma2025sql}.

\section{Conclusion}
We propose Graph-Reward-SQL, a reward model framework for RL-based Text-to-SQL that replaces execution-based rewards with the GMNScore outcome reward model and the StepRTM stepwise reward model. By eliminating the dependency on execution during training, it significantly improves training efficiency. Extensive experiments demonstrate that our proposed reward models achieve superior performance. The Graph-Reward-SQL framework establishes a new direction toward scalable and efficient RL-based Text-to-SQL tasks.

\section{Limitations}\label{appendix:limitation}
While our proposed reward models demonstrate strong performance in text-to-SQL tasks and show much better performance than the execution-based reward model, its SQL-specific architectural design currently restricts its applicability to broader code generation domains. Generalizing our reward model framework to support programming languages such as Python and Java would require some modifications to the language representation mechanisms, particularly to address the challenge of assessing functional equivalence across diverse programming paradigms.

However, this limitation opens promising directions for future work. We aim to explore the applicability of the execution-free reward model in broader code generation tasks in future work, thereby offering viable alternatives to the common RL practice of relying solely on execution outcomes. This extension may contribute to more general, efficient, and flexible reinforcement learning training for code generation tasks.
\bibliography{custom}

\appendix

\label{sec:appendix}

\clearpage

\section{Datasets Details}
\label{appendix:datasets}
\paragraph{\bf Spider} \citep{yu2018spider} is a large-scale, cross-domain Text-to-SQL benchmark containing 10,181 questions and 5,693 complex SQL queries. These queries come from 200 multi-table databases across 138 domains. The dataset is split into three subsets without any overlap in databases: training (8659), development (1034) and test datasets. This separation helps ensure a fair assessment of model performance.

\paragraph{\bf BIRD} \citep{li2024can} is another large-scale and cross-domain Text-to-SQL dataset known for its complexity. It contains 12,751 question-SQL pairs drawn from 95 databases across more than 37 professional fields, making it a challenging testbed for evaluating the generalization capability of semantic parsers. A notable characteristic of BIRD is its “dirty” nature: queries may be inaccurate, column names are often ambiguous or poorly described, and databases may contain null values or irregular encodings. These issues collectively pose unique challenges for model robustness. The dataset is split into 9,428 training examples and 1,534 development examples, with the remaining reserved for testing.

\paragraph{\bf 200k-Text2SQL} comprises 200,000 examples related to the text-to-SQL task. Each data entry also includes a question, a database schema, and a reference SQL statement. To prepare the model for generating clean SQL queries during the PPO phase, we first fine-tune the Deepseek-coder-1.3b-ins model via Supervised Fine-Tuning (SFT), which enables the model to follow instructions and complete Text-to-SQL tasks effectively. The prompt used is detailed in \ref{appendix:prompt strategie}. The dataset is available at: \url{https://huggingface.co/datasets/philikai/200k-Text2SQL}.

\section{Evaluation Setting Details}\label{appendix:evaluation_setting_details}
Test-Suite (TS) Performance provides a robust and reliable assessment of semantic accuracy. For example, the original \texttt{academic.sqlite} database is automatically augmented into 411 variants, resulting in a total of 412 test databases. Each predicted-reference SQL pair is executed on all of them. This reduces the likelihood of false positives and strengthens the evaluation's ability to detect semantic discrepancies.

In our evaluation using the official TS framework, we emphasize strictness in semantic verification. Specifically, we enable the \texttt{--keep\_distinct} flag to preserve the use of the \texttt{DISTINCT} keyword during evaluation. Meanwhile, we disable \texttt{--plug\_value} to avoid injecting ground-truth values into predicted queries, thus requiring the model to generate complete SQL outputs, including correct values. We use the TS code to evaluate Spider on its augmented dev split databases. For BIRD, since no augmented databases are provided, we apply the TS code directly on its original database. The code and augmented databases are available at: \url{https://github.com/taoyds/test-suite-sql-eval}.

In our experiments using the TS evaluation metric, all hints (e.g., age = year - birth\_year) in BIRD dataset were deliberately omitted to align with the setup of the Spider dataset. This methodological choice results in lower performance metrics compared to studies that incorporated the hints available in the BIRD dataset.

\section{Implement Details}\label{appendix:hyperparameter_setting}
\subsection{Outcome Reward Experiments} 

For the outcome reward PPO experiments, we first conduct a warm-up phase using a different dataset from the target training corpus—200k-Text2SQL. We sample an equivalent number of examples as in the training split of Spider dataset and perform 2 epochs of SFT warm-up using SFT. The sampling process also involves removing irrelevant samples, such as those with queries in Chinese, to ensure consistency and relevance in the training data. This warm-up helps the model adapt to prompt-following behavior and prevents it from generating non-SQL text. After SFT warm-up, PPO training proceeds on the Spider/BIRD dataset with hyperparameters shown in table~\ref{tab:ppo hyper}. These hyperparameters, along with those for our reward model, were either adopted from prior work \citep{shojaee2023execution} or determined through empirical experiments.

\begin{table}[h]
\centering
\caption{PPO hyperparameters.}
\resizebox{1\linewidth}{!}{%
\begin{tabular}{ll} 
\toprule
\textbf{Hyperparameter} & \textbf{Default Value} \\
\midrule
Optimizer & AdamW \\
Learning Rate & 5e-6 \\
Scheduler & Constant \\
Max Length & 1024 \\
Batch Size & 32 \\
$N_\text{mb}$ Number of Mini-batches & 8 \\
Gradient Accumulation Steps & 8 \\
$\beta$ (KL Penalty Coefficient) & 0.05 \\
$\gamma$ (Discount Factor) & 0.99 \\ 
$\lambda$ (for GAE) & 0.99 \\ 
$K$ (PPO Update Iteration)& 1 \\
$\varepsilon$ (PPO's Policy Clipping Coefficient) & 0.1 \\
Value Function Loss Clipping
& True\\
$\hat{\varepsilon}$ (Value Clipping Coefficient) & 0.2 \\
Sampling Temperature & 1 \\
\bottomrule
\end{tabular}%
}
\label{tab:ppo hyper}
\end{table}

\subsection{Stepwise Reward Experiments} 
For the stepwise reward PPO experiments, we encourage the generation of CTE-structured SQLs by incorporating CTE-rewritten data into the SFT stage. Since SQL queries in BIRD are generally more complex than those in Spider, they are particularly well-suited for decomposition into modular subqueries using the CTE format. To leverage this, we rewrite the reference SQLs in the BIRD training set into CTE format using GPT-4o, following the prompt strategies outlined in Appendix~\ref{appendix:prompt strategie}. The resulting data is then combined with the warm-up dataset used in the outcome reward model experiments to construct a SFT warm-up set. This setup encourages the model to generate CTE SQL, enabling effective application of stepwise reward signals during PPO training.

\subsection{Qwen2.5-Coder-7B-Ins Experiments}
\label{appendix:GRPO_setup}
Given the stronger capabilities of the base model, we directly applied reinforcement learning to the Qwen2.5-Coder-7B-Ins model using two distinct RL protocols, GRPO and PPO, without any warm-up training.

In experiments of Group Relative Policy Optimization (GRPO) with Qwen2.5-Coder-7B-Ins, we configured the learning rates for the actor and critic to 5e-6 and 2e-6, respectively, with a warm-up period of 10 steps to ensure stable training dynamics in the early stages. The model was trained with a batch size of 512, and mini-batches of size 256 were used. A weight decay of 0.1 was applied to regularize the model and mitigate overfitting. The group size for GRPO was set to 16, which corresponds to the number of completions generated per input prompt, thereby facilitating the optimization of the policy through relative comparisons between groups. For effective handling of long sequences, we set the maximum prompt length to 4096 and the maximum response length to 16,384.

\begin{figure}[t!]
\centering
\includegraphics[width=1\linewidth]{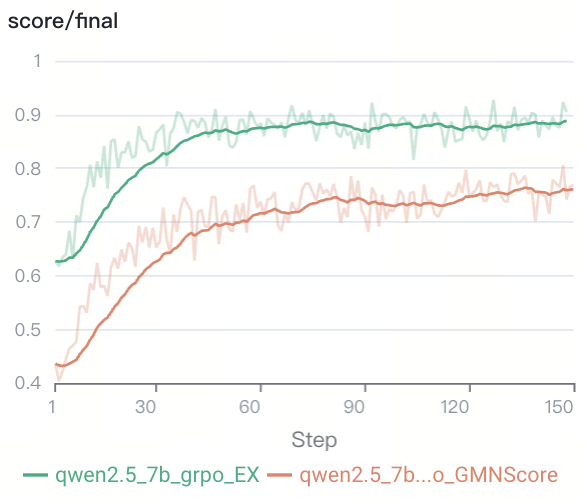}
\vspace{-15pt}
\caption{Reward score trends on the training set during GRPO training. The green curve corresponds to the execution-based reward model (EX), while the orange curve shows the results using the proposed reward model GMNScore. Both models are trained with the Qwen2.5-Coder-7B-Ins backbone under the GRPO framework. While GMNScore starts from a lower initial score compared to EX, it demonstrates a larger overall increase.}
\label{appendix:fig:training_reward}
\vspace{-10pt}
\end{figure}

\begin{figure}[t!]
\centering
\includegraphics[width=1\linewidth]{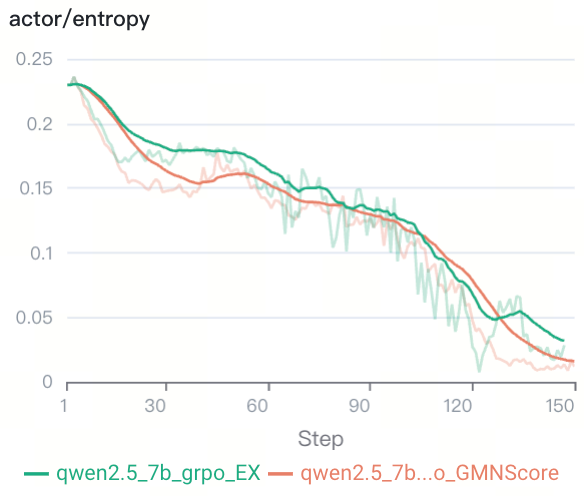}
\vspace{-15pt}
\caption{Actor entropy of GRPO training. Both models are based on Qwen2.5-Coder-7B-Ins. The orange curve corresponds to GMNScore and the green curve to EX. Entropy consistently decreases as the policy becomes more deterministic. GMNScore exhibits a faster reduction in entropy during the early training stages, while both methods converge to similarly low-entropy levels by the end of training.}
\vspace{-10pt}
\label{appendix:fig:training_entropy}
\end{figure}

We visualized the reward scores and entropy values of the training dataset during the training process, as shown in Figure~\ref{appendix:fig:training_reward} and Figure~\ref{appendix:fig:training_entropy}.

\section{GRPO}
\label{appendix:GRPO}
\noindent
In addition to experiments based on PPO, we also evaluate our methods under the Group Relative Policy Optimization (GRPO) RL framework~\cite{shao2024deepseekmath}, in order to further assess the effectiveness of our proposed GMNScore outcome reward model. In the standard text-to-SQL setting, let $x$ denotes a natural language query, and $\mathcal{G} = \{\hat{q}_1, \dots, \hat{q}_G\}$ denotes a group of $G$ candidate SQL queries sampled from the current policy $\pi_\theta(\cdot | x)$. To better leverage relative preferences among candidates, GRPO optimizes the policy by maximizing the following objective:
$$
\begin{aligned}
\mathcal{J}_{\text{}}(\theta) = &\mathbb{E}_{x \sim \mathcal{D},\, \mathcal{G} \sim \pi_\theta(\cdot|x)^G} \\
\bigg[ 
& \frac{1}{G} \sum_{i=1}^{G} \min \left( \frac{\pi_\theta(\hat{q}_i | x)}{\pi_{\theta_{\mathrm{old}}}(\hat{q}_i | x)} A_i,\; \right. \notag \\
& \quad\left. \mathrm{clip}\left( \frac{\pi_\theta(\hat{q}_i | x)}{\pi_{\theta_{\mathrm{old}}}(\hat{q}_i | x)}, 1 - \epsilon, 1 + \epsilon \right) A_i \right) \notag \\
& - \beta \, \mathbb{D}_{\mathrm{KL}}\left(\pi_\theta(\cdot \mid x)\,\|\,\pi_{\mathrm{ref}}(\cdot \mid x)\right) \bigg],
\end{aligned}
$$

\noindent where $\pi_{\theta}$ and $\pi_{\theta_{\mathrm{old}}}$ denote the current and previous policies, respectively, and $\pi_{\mathrm{ref}}$ is a fixed reference policy used for KL regularization. The term $A_i$ represents the advantage of candidate $\hat{q}_i$ within the sampled group, reflecting its relative quality compared to other completions. The ratio $\frac{\pi_\theta(\hat{q}_i | x)}{\pi_{\theta_{\mathrm{old}}}(\hat{q}_i | x)}$ captures the likelihood shift under the current policy, while $\epsilon$ and $\beta$ are hyperparameters that control the clipping threshold and the strength of the KL penalty, respectively. 

By generating multiple candidates per input, GRPO naturally accommodates the inherent ambiguities and challenges of mapping natural language to SQL queries, ensuring that feedback is both robust and informative.

\section{Training of GMN of GMNscore}
\label{appendix:gmn_train}
The Graph Matching Network (GMN) proposed serves as a foundation for assessing SQL query equivalence \citep{zhan-etal-2025-towards}, it was not originally designed for RL-based training. In our preliminary evaluation, reward model GMNScore built on the provided GMN model of FuncEvalGMN \citep{zhan-etal-2025-towards} lagged behind that of the baseline EX under PPO experiments. We conducted a failure case analysis focusing on example where scores of GMNScore reward model disagreed with execution results. After manually checking, we identified several types of potential error situations.

To develop a GMN better suited for RL-based Text-to-SQL tasks, we retrain the model using an augmented version of the Spider-train-pair dataset introduced in \citep{zhan-etal-2025-towards}. The original Spider-train-pair dataset consists of 17,664 SQL query pairs annotated with binary equivalence labels. Two corresponding sets for evaluation, Spider-dev-pair and BIRD-dev-pair, contain 1,644 and 2,977 examples, respectively.

Seven augmentation strategies were crafted from the identified failure cases, adding 3,397 training pairs. The strategies are summarized in Table~\ref{tab:sql_augmentation_examples}. After augmentation, the training set exceeded 20,000 SQL pairs.

\begin{table*}[ht!]
\centering
\scriptsize
\resizebox{\textwidth}{!}{
\begin{tabular}{c|p{0.2\linewidth}|p{0.35\linewidth}|p{0.35\linewidth}}
\toprule
\textbf{Number} & \textbf{Augmentation Type} & \textbf{Original SQL} & \textbf{Augmented Result} \\
\midrule
\multicolumn{4}{c}{\textbf{Enhancement of equivalent cases}} \\
\midrule
137 & \parbox[t]{\linewidth}{
IN Clause Replacement
} 
& \parbox[t]{\linewidth}{\ttfamily
\textbf{Original:} SELECT Name FROM (SELECT Name, Age FROM technician) AS t \\
\sethlcolor{LightGreen}\hl{WHERE Age IN (36, 37)}
}
& \parbox[t]{\linewidth}{\ttfamily
\textbf{Aug:} SELECT Name FROM (SELECT Name, Age FROM technician) AS t \\
\sethlcolor{LightGreen}\hl{WHERE Age = 36 OR Age = 37}
}
\\
\midrule
\multicolumn{4}{c}{\textbf{Enhancement of non-equivalent cases}} \\
\midrule
422 & \parbox[t]{\linewidth}{
Column Name Perturbation \\ (Select Clause)
} 
& \parbox[t]{\linewidth}{\ttfamily
\textbf{Original:} SELECT \sethlcolor{LightRed}\hl{actid} FROM activity \\ }
&
\parbox[t]{\linewidth}{\ttfamily
\textbf{Aug:} SELECT \sethlcolor{LightRed}\hl{acti} FROM activity \\
\textbf{Aug:} SELECT \sethlcolor{LightRed}\hl{ctid} FROM activity \\ }
\\
\midrule
566 & 
\parbox[t]{\linewidth}{
Keyword Replacement \\ (AND/OR)
} 
 & 
\parbox[t]{\linewidth}{\ttfamily
\textbf{Original:} SELECT * FROM Products \\
WHERE Price >= 60 \sethlcolor{LightRed}\hl{AND} Price <= 120
}
& 
\parbox[t]{\linewidth}{\ttfamily
\textbf{Aug:} SELECT * FROM Products \\
WHERE Price >= 60 \sethlcolor{LightRed}\hl{OR} Price <= 120 \\
}
\\\midrule
1094 & \parbox[t]{\linewidth}{
Symbol Replacement \\ (Comparison Operator)
} 
& \parbox[t]{\linewidth}{\ttfamily
\textbf{Original:} SELECT * FROM (SELECT dept\_name, building FROM department) AS t \\
WHERE building \sethlcolor{LightRed}\hl{>} (SELECT AVG(building) FROM department)
}
& \parbox[t]{\linewidth}{\ttfamily
\textbf{Aug:} SELECT * FROM (SELECT dept\_name, building FROM department) AS t \\
WHERE building \sethlcolor{LightRed}\hl{>=} (SELECT AVG(building) FROM department)
}
\\
\midrule
276 & \parbox[t]{\linewidth}{
Table Source Replacement
} 
& \parbox[t]{\linewidth}{\ttfamily
\textbf{Original:} SELECT * FROM (SELECT name, email FROM \sethlcolor{LightRed}\hl{user\_profiles}) AS t \\
WHERE name LIKE '\%Swift\%'
}
& \parbox[t]{\linewidth}{\ttfamily
\textbf{Aug:} SELECT * FROM (SELECT name, email FROM \sethlcolor{LightRed}\hl{user}) AS t \\
WHERE name LIKE '\%Swift\%'
}
\\
\midrule
64 & \parbox[t]{\linewidth}{
Column Name Replacement
} 
& \parbox[t]{\linewidth}{\ttfamily
\textbf{Original:} SELECT \sethlcolor{LightRed}\hl{candidate\_id} FROM candidate \\
ORDER BY candidate\_id DESC LIMIT 3
}
& \parbox[t]{\linewidth}{\ttfamily
\textbf{Aug:} SELECT \sethlcolor{LightRed}\hl{people\_id} FROM candidate \\
ORDER BY people\_id DESC LIMIT 3
}
\\
\midrule
838 & \parbox[t]{\linewidth}{
Column Removal
} 
& \parbox[t]{\linewidth}{\ttfamily
\textbf{Original:} SELECT \sethlcolor{LightRed}\hl{circuitid, location} FROM (SELECT circuitid, location, country FROM circuits) AS t \\
WHERE country = 'fraNce' OR country = 'belGium'
}
& \parbox[t]{\linewidth}{\ttfamily
\textbf{Aug:} SELECT \sethlcolor{LightRed}\hl{circuitid} FROM (SELECT circuitid, country FROM circuits) AS t \\
WHERE country = 'fraNce' OR country = 'belGium'
}
\\
\bottomrule
\end{tabular}
}
\caption{Examples of SQL augmentation strategies used for generating non-equivalent/equivalent SQL pairs. Each example introduces specific perturbations (e.g., column name change or logical operator replacement) that alter the semantics of the original SQL.}
\label{tab:sql_augmentation_examples}
\end{table*}

\section{Details of Execution Accuracy (EX)}
\label{appendix:ex}
Execution Accuracy (EX) is a commonly used reward signal in text-to-SQL tasks, offering supervision based on the correctness of query execution. However, we do not adopt the naive binary EX metric that only distinguishes between success and failure. Instead, we adopt a stronger variant of execution-based reward that integrates syntax check signals and runtime diagnostics, resulting in a more fine-grained supervision signal. Motivated by prior work~\cite{pourreza2025reasoning, shojaee2023execution}, this reward formulation extends traditional binary execution accuracy with syntax awareness and serves as a strong execution-based baseline for comparison.

Given a generated SQL query $\hat{q}$ and the reference SQL $q^\star$, we compute the EX reward as:
$$
r_{\text{EX}}(\hat{q},q^\star) = \begin{cases}
1, & \text{correct execution} \\
-0.3, & \text{incorrect execution} \\
-0.6, & \text{runtime error} \\
-1, & \text{syntax error}
\end{cases}
$$

A runtime error refers to a situation where the SQL query is syntactically correct but fails during execution. This can occur, for example, when a non-existent table or column is referenced in the query. This reward formulation provides more informative feedback to the policy model, especially during early training stages when most queries fail due to syntax or runtime errors, avoiding undifferentiated negative signals.

For the false positive problem of execution-based rewards, Test Suite (TS)~\citep{zhong2020semantic} attempted to improve robustness by using a set of test cases to simulate query behavior under different data distributions. Nevertheless, as shown in \citep{zhan-etal-2025-towards}, both false positives (semantically incorrect queries that return the right result) and false negatives (semantically correct queries that fail under test data) persist, due to the reliance on incomplete or ambiguous database contents.

Furthermore, frequent SQL execution significantly increases computational overhead, becoming the major bottleneck during RL training, where query evaluation must occur at each rollout step.

\section{Cost Analysis of Reward Models}
\label{appendix:cost}
To evaluate the computational efficiency of different reward models, we conduct experiments on an edge server equipped with an Intel Xeon Platinum 8336C CPU (128 cores) and a total memory capacity of 2.0~TiB. Our comparison focuses exclusively on model-based reward functions. Rule-based reward methods (AstPM/RelPM outcome reward model and our proposed stepwise reward model StepRTM) are excluded from this analysis as they incur negligible memory overhead.

\begin{table}[h]
\centering
\small
\begin{tabular}{lccc}
\toprule
\textbf{Reward Type} & \textbf{Time$\downarrow$} & \textbf{Params$\downarrow$} & \textbf{GPU Memory$\downarrow$} \\
\midrule

EX & 1.088s & N/A & N/A \\
BTRM & 0.095s & 1.35B & 9304MB \\
GMNScore & \textbf{0.023s} & \textbf{3.99M} & \textbf{83MB} \\
\bottomrule
\end{tabular}
\caption{Cost comparison across reward models.}
\label{tab:reward_cost}
\end{table}

\paragraph{Time Cost.} 
We use the training split of the Spider dataset and perform one epoch of PPO training, during which we measure the total time consumed by the reward score computation. We then record the average reward calculation time per sample, providing insights into the computational efficiency of different reward models. Both BTRM and GMNScore are implemented using bfloat16 precision for acceleration. Execution-based reward (EX) incurs the highest computational cost due to repeated database calls. While EX can leverage high concurrency when executed on the CPU to accelerate performance, it still remains slower in practice due to its reliance on database calls. In real-world scenarios, different GPU rollouts of samples are processed concurrently through the reward model, which can improve processing speed. However, both BTRM and GMNScore can also easily achieve parallelization by loading the same model onto each GPU. Therefore, the speed reflected in the table is representative of the actual performance in training. 

\paragraph{Model Size.}
We also compare the model sizes based on the total number of parameters using:
\begin{lstlisting}[language=Python, basicstyle=\ttfamily\small]
p.numel() for p in model.parameters()
\end{lstlisting}
We train Bradley-Terry Reward Model (BTRM) using DeepSeek-Coder-1.3B-Ins as the base model, with a total of 1,346,471,936 parameters. In contrast, GMNScore adopts a lightweight Graph Matching Network with only 3,994,944 parameters.

\paragraph{Memory Cost.}
To evaluate GPU memory consumption during inference, we apply:
\begin{lstlisting}[language=Python, basicstyle=\ttfamily\small]
torch.cuda.reset_peak_memory_stats()
# here is model inference code
torch.cuda.max_memory_allocated()
\end{lstlisting}
On average, BTRM consumes approximately 9304~MB of GPU memory, while GMNScore requires only 83~MB.

\section{Analysis of GMNScore Accuracy} 
\label{sec:gmn_better_rlhf_analysis}

\begin{figure}[t!]
\centering
\includegraphics[width=0.9\linewidth]{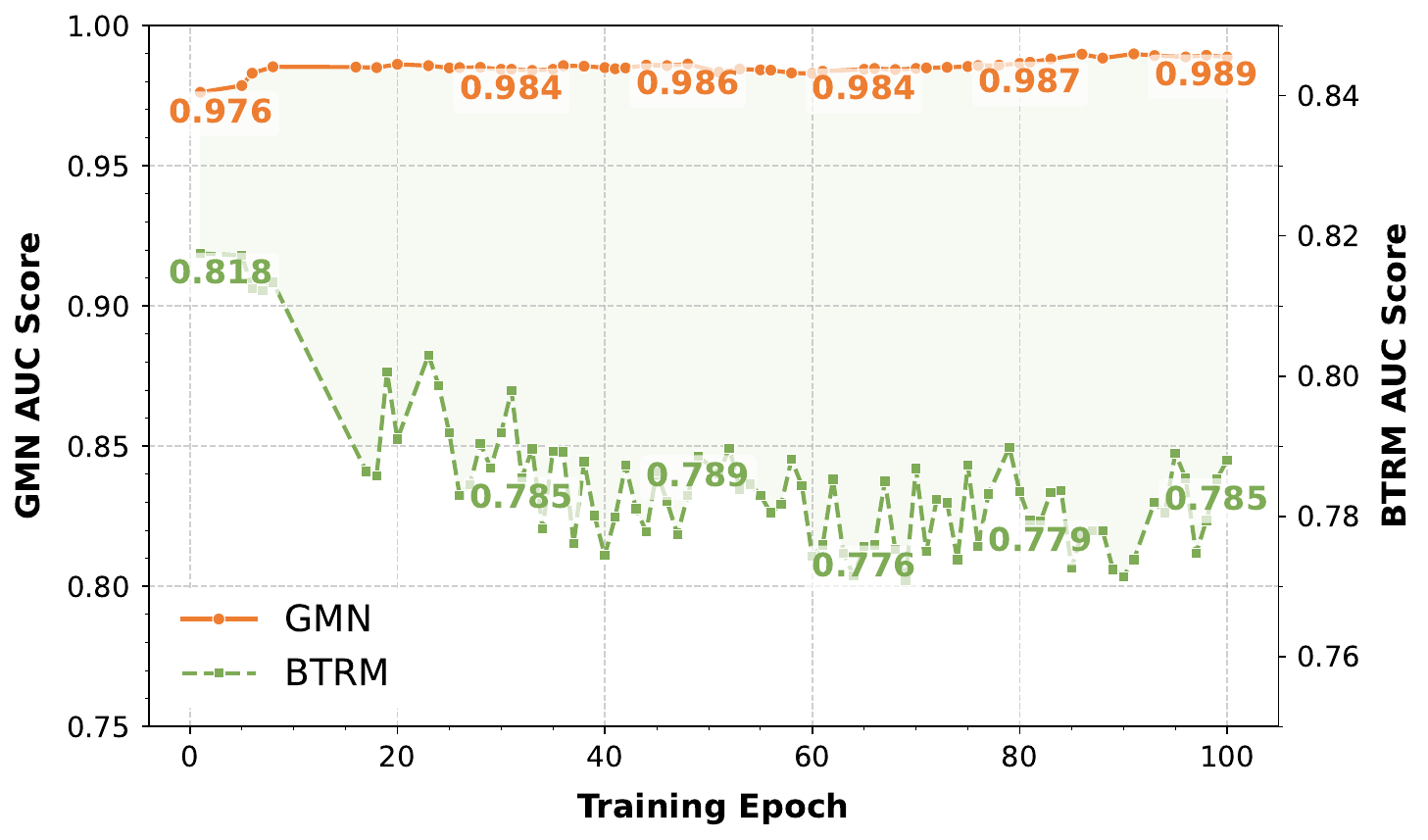}
\vspace{-5pt}
\caption{There is a high Area Under the Curve (AUC) score between the GMNScore and execution results, consistently exceeding 97.6\% during training.}
\label{fig:performance_comparison_gmn_llm}
\vspace{-10pt}
\end{figure}

Experimental results demonstrate the effectiveness of GMNScore as a reward model in PPO, significantly outperforming BTRM. We analyze the correlation\footnote{AUC is used as a metric to assess the accuracy of similarity evaluation \cite{zhan-etal-2025-towards}. In our experiments, a score of 100\% does not indicate optimal performance, as false negatives are present in the execution outcomes.} between these two reward signals and actual execution outcomes during PPO training. As shown in Figure~\ref{fig:performance_comparison_gmn_llm}, GMNScore consistently maintains a high correlation with the execution results. This indicates that GMNScore provides a more stable and precise reward signal than BTRM during training, contributing to its superior performance.

\section{Hybrid Outcome Reward Designs}
\label{appendix:hybrid_outcome}
\begin{table}[ht!]
\centering
\scriptsize
\renewcommand{\arraystretch}{1.2}
\resizebox{0.45\textwidth}{!}{
\begin{tabular}{llccc}
\toprule
\textbf{Method} & \textbf{Spider} & \textbf{BIRD} & \textbf{Avg.} \\
\midrule
\textbf{DeepSeek-Coder-1.3B-Ins} & 39.56 & 11.34 & 25.45 \\
 + PPO w/ RelPM & 62.86 & 14.67 & 38.77 \\
 + PPO w/ EX & 65.28 & 17.21 & 41.25 \\
 + PPO w/ GMNScore & 67.70 & 16.10 & 41.90 \\
\midrule
+ PPO w/ EX \& AstPM & 62.38 & 16.43 & 39.41 \\
+ PPO w/ GMNScore  \& AstPM & 65.28 & 16.17 & 40.73 \\
+ PPO w/ GMNScore  \& RelPM & 66.34 & 16.69 & 41.52 \\
+ PPO w/ EX \& GMNScore  & 67.02 & 17.73 & 42.38 \\
+ PPO w/ EX \& RelPM & \textbf{68.28} & \textbf{18.77} & \textbf{43.53} \\
\bottomrule
\end{tabular}
}
\caption{TS Performance of DeepSeek-Coder-1.3B-Ins models under different combinations of outcome reward strategies.}
\label{tab:outcome_combination}
\end{table}

We investigate whether combining multiple outcome reward models can further improve performance. Inspired by recent work that integrates multiple reward signals \citep{pourreza2025reasoning, ma2025sql}, we explore various combinations among EX, GMNScore, RelPM, and AstPM. As shown in Table~\ref{tab:outcome_combination}, combining EX with RelPM achieves the best overall performance, increasing the average accuracy from 41.90\% (GMNScore alone) to 43.53\%. The second-best result comes from EX combined with GMNScore, reaching 42.38\%.

We attribute the relatively lower performance of EX \& GMNScore compared to EX \& RelPM to the fact that both EX and GMNScore emphasize global correctness. This may lead to redundant or even conflicting feedback, limiting the effectiveness of their combination. In contrast, the integration of RelPM provides localized partial matching signals, which offer complementary supervision and improve the performance.

Moreover, combining AstPM with either GMNScore or EX does not outperform using GMNScore or EX alone. We suspect this is due to AstPM's focus on surface-level syntax. It may penalize syntactically different yet semantically equivalent SQLs, introducing false negatives that degrade training quality. As a result, this syntactic noise may undermine the semantic robustness provided by GMNScore or EX.

While GMNScore \& RelPM performs better than RelPM alone, it still falls short of GMNScore alone. We believe this is because RelPM, although based on ROT to capture richer structural semantics, still relies on partial matching algorithm, which struggles to fully capture deep semantic equivalence. By contrast, GMNScore directly leverages graph matching to assess functional equivalence, demonstrating strong robustness and stability.

It is worth noting, however, that the effectiveness of combining multiple outcome rewards is significantly lower than that of our proposed Stepwise Reward model, StepRTM. The highest value achieved by the outcome rewards combination is 43.53\%, whereas StepRTM achieves a higher value of \textbf{45.32\%}, representing an improvement of \textbf{1.79\%}. This highlights the effectiveness of our designed stepwise reward model StepRTM. More importantly, StepRTM introduces no additional costs in terms of database execution time or GPU memory consumption.

\section{Generalization of GMNScore Reward}
\begin{table}[h]
\centering
\scriptsize
\renewcommand{\arraystretch}{1.2}
\resizebox{0.45\textwidth}{!}{
\begin{tabular}{llccc}
\toprule
\textbf{Method} & \textbf{Spider} & \textbf{BIRD} & \textbf{Avg.} \\
\midrule
\textbf{DeepSeek-Coder-1.3B-Ins} & 39.56 & 11.34 & 25.45 \\
+ PPO w/ AstPM & 51.26 & 11.47 & 31.37 \\
+ PPO w/ RelPM & 51.45 & 14.73 & 33.09 \\
+ PPO w/ BTRM & 51.35 & 15.45 & 33.40 \\
+ PPO w/ EX & 52.71 & \textbf{15.84} & 34.28\\
+ PPO w/ GMNScore (Our) & \textbf{53.87} & 15.58 & \textbf{34.73} \\
\bottomrule
\end{tabular}
}
\caption{TS Performance of DeepSeek-Coder-1.3B-Ins models under different outcome rewards with BIRD-Train.}
\label{tab:bird_results}
\end{table}

Although the GMN of GMNScore reward model is trained solely on SQL equivalence data derived from the Spider dataset, it exhibits strong cross-database generalization without requiring any additional fine-tuning. As shown in Table~\ref{tab:bird_results}, when RL is conducted on the training split of BIRD, GMNScore continues to outperform execution-guided (EX) reward in average TS performance, consistent with the results on training split of Spider.

More broadly, we observe that across both training configurations of Spider and BIRD, our GMNScore consistently achieves higher average validation performance than EX. However, a finer-grained comparison reveals a nuanced trend: on both DeepSeek-Coder-1.3B and 6.7B with PPO, GMNScore consistently outperforms EX on the Spider evaluation dataset but shows a slight performance drop on the BIRD evaluation dataset.

We attribute this discrepancy to the inherent limitations of the GMN model itself. Although we enhanced GMN through data augmentation based on the Spider-train-pair dataset \cite{zhan-etal-2025-towards}, the augmented data is still constructed from Spider-style SQL, which biases the model toward better alignment with the structural and stylistic patterns of Spider. In contrast, the BIRD dataset is significantly more challenging, containing more complex SQL constructs such as deeply nested subqueries. As a result, GMN may struggle to judge equivalence in BIRD-style queries as accurately as its performance in Spider. For example, GMN is likely to score generated SQL in Spider-like styles more accurately, leading to more consistent reinforcement signals, while its scoring fidelity for BIRD-like styles is relatively weaker.

In contrast, the EX reward is purely execution-based and thus remains more robust to variations in dataset complexity and style. However, EX also suffers from false positives. Despite this slight drop in the BIRD dataset, GMNScore offers substantially larger improvements on Spider and achieves the best overall average performance. We further hypothesize that stronger base models can compensate for the small drop in reward signal quality in the BIRD dataset.

As discussed previously, GMNScore exhibits slightly reduced BIRD performance compared to EX when using Deepseek-Coder-1.3b/6.7b-Ins models, likely due to GMN’s limited generalization on BIBD-style SQL. However, this performance gap decreases with stronger base models. As shown in Table~\ref{tab:grpo_result}, when using Qwen-Coder-7B and 14B as the policy backbone, GMNScore not only consistently outperforms EX on Spider, but also achieves superior results on BIRD. This further supports the effectiveness and scalability of our reward model framework across different model capacities.

\begin{figure*}[ht]
    \centering
    \includegraphics[width=1\linewidth]{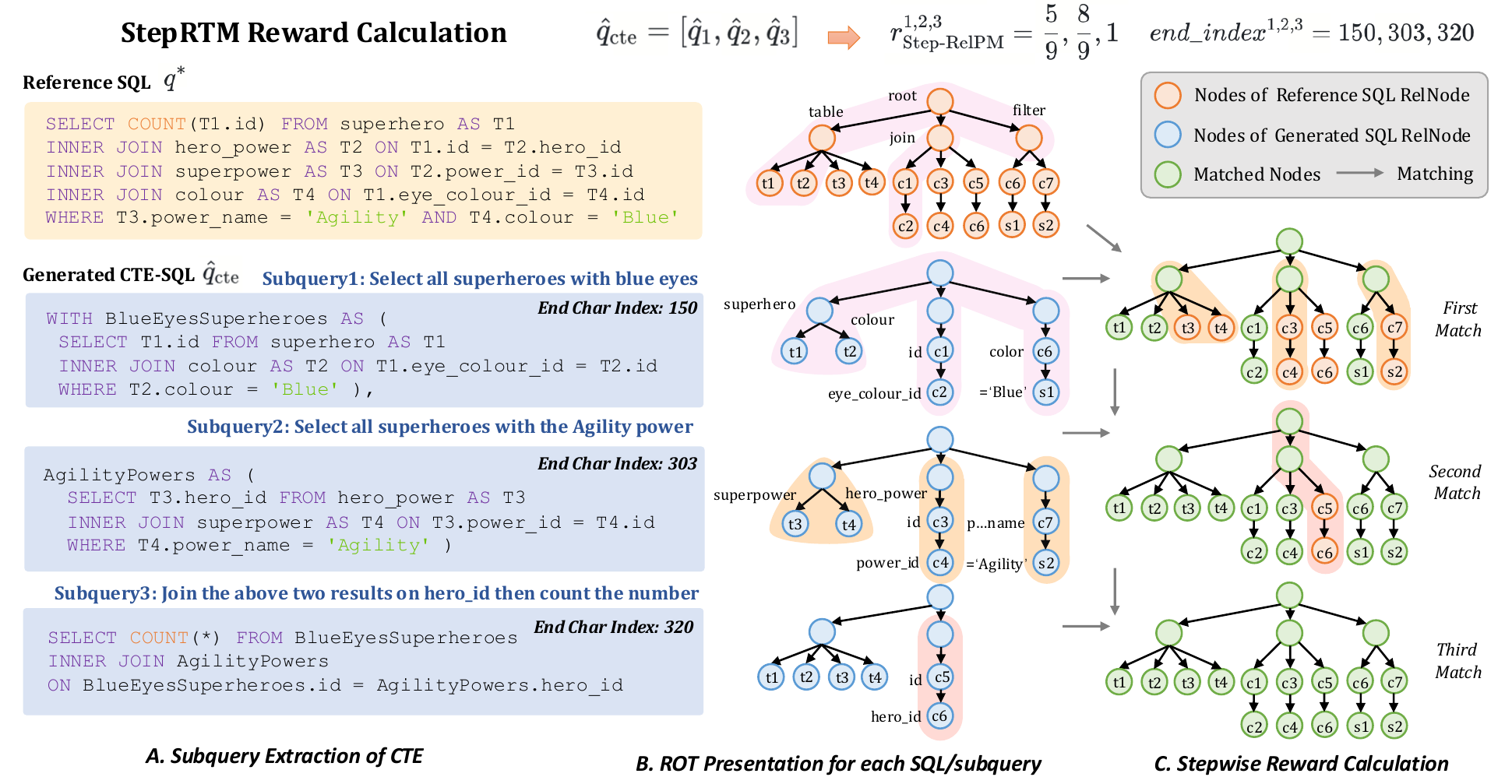}
    \caption{Detailed Overview of the StepRTM Stepwise Reward Calculation.}
    \label{fig:step_relpm_pipeline_detail}
\vspace{-10pt}
\end{figure*}

\section{Detailed Effectiveness of GMNScore Across GRPO}
\label{appendix:exp_grpo_result}
\vspace{-5pt}
\begin{table}[h!]
\centering
\scriptsize
\renewcommand{\arraystretch}{1.2}
\resizebox{0.45\textwidth}{!}{
\begin{tabular}{llccc}
\toprule
\textbf{Method} & \textbf{Spider} & \textbf{BIRD} & \textbf{Avg.} \\
\midrule
\textbf{Qwen2.5-Coder-7B-Ins}  & 62.67 & 22.69 & 42.68 \\
+ PPO w/ EX & 75.24 & 29.01 & 52.13 \\
+ PPO w/ GMNScore (Ours) & \textbf{76.89}\textcolor{red!75!black}{\tiny $\uparrow$1.65} & \textbf{29.60}\textcolor{red!75!black}{\tiny $\uparrow$0.59} & \textbf{53.25}\textcolor{red!75!black}{\tiny $\uparrow$1.12} \\
+ GRPO w/ EX & 76.40 & 28.49 & 52.45 \\
+ GRPO w/ GMNScore (Ours) & \textbf{78.53}\textcolor{red!75!black}{\tiny $\uparrow$2.13} & \textbf{29.92}\textcolor{red!75!black}{\tiny $\uparrow$1.43} & \textbf{54.23}\textcolor{red!75!black}{\tiny $\uparrow$1.78} \\
\midrule
\textbf{Qwen2.5-Coder-14B-Ins}  & 71.08& 29.29& 50.19 \\
+ PPO w/ EX & 74.66 & 32.98 & 53.82 \\
+ PPO w/ GMNScore (Ours) & \textbf{77.37}\textcolor{red!75!black}{\tiny $\uparrow$2.71} & \textbf{33.70}\textcolor{red!75!black}{\tiny $\uparrow$0.72} & \textbf{55.54}\textcolor{red!75!black}{\tiny $\uparrow$1.72} \\
+ GRPO w/ EX & 77.56 & 33.90 & 55.73 \\
+ GRPO w/ GMNScore (Ours) & \textbf{78.34}\textcolor{red!75!black}{\tiny $\uparrow$0.78} & \textbf{34.35}\textcolor{red!75!black}{\tiny $\uparrow$0.45} & \textbf{56.35}\textcolor{red!75!black}{\tiny $\uparrow$0.62} \\

\bottomrule
\end{tabular}
}
\caption{TS Performance of Qwen2.5-Coder-7B/14B-Ins models directly trained by PPO/GRPO under EX and GMNScore outcome rewards.}
\label{tab:grpo_result}
\end{table}

As shown in Table~\ref{tab:grpo_result}, our proposed GMNScore consistently outperforms the EX across both PPO and GRPO training paradigms, and for both Qwen2.5-Coder-7B-Ins and 14B-Ins models. On the 7B model, GMNScore improves the average score from 52.13\% to 53.25\% under PPO training (\textbf{+1.12\%}), and from 52.45\% to 54.23\% under GRPO training (\textbf{+1.78\%}). Similarly, on the 14B model, GMNScore boosts the average score by \textbf{+1.72\%} under PPO and \textbf{+0.62\%} under GRPO. These consistent improvements demonstrate that GMNScore provides a stable reward signal than EX. Overall, these results confirm the superiority of GMNScore in guiding RL-based Text-to-SQL tasks more effectively than execution-based EX reward model.

\section{StepRTM Calculation}\label{appendix:StepRTM}
In Figure~\ref{fig:step_relpm_pipeline} in Methodology part, we presented an extremely simplified example. As shown in Figure~\ref{fig:step_relpm_pipeline_detail}, we offer a figure with more information to help better understand the StepRTM calculation process.

\section{Training Details of Neural Reward Model with Language odeling}\label{appendix:RLHF_LLM_training}

\paragraph{\bf Preference Data Construction} We construct preference data using the \textbf{Spider-Train Pair} from GMN’s training set. Preference pairs are formed through label-driven selection, where chosen responses strictly correspond to \(\text{label} = 1\), mapped from the \texttt{new\_labels} column, while rejected responses (negative samples) come from the same prompt cluster. Within each group, all possible pairs between chosen (\(\text{label} = 1\)) and rejected (\(\text{label} \neq 1\)) responses are generated.

\paragraph{\bf Reward Model Training} To train the reward model in the above preference pairs, we finetune the pretrained model DeepSeek-Coder-1.3B-Instruct with a reward head. Training is carried out using an effective batch size of \(16\) with gradient accumulation \(2\). We use the AdamW optimizer, a learning rate of \(1.41 \times 10^{-5}\), and a linear warm-up schedule. The model is trained for five epochs with a sequence length of 2048 tokens. 

\paragraph{Prompt Engineering Strategy}
The reward model should effectively evaluates the generated SQL based on both syntactic clarity and execution-level correctness. To reinforce schema understanding and functional equivalence, the data was structured as follows:
\begin{tcolorbox}[colframe=black!20!white, colback=white, coltitle=black, title=\textbf{Dataset}]
\begin{verbatim}
[Table Schema] 
[Question] 
[Reference SQL] 
Please give a SQL that is functionally
identical to the SQL above:
[Chosen SQL/Rejected SQL] 
\end{verbatim}
\end{tcolorbox}

\section{RelNode-based Partial Matching}\label{appendix:relpm_detail}
To enable semantic-level comparison of SQL queries, each query is converted into a Relational Operator Tree (ROT)—a representation of the query's logical execution plan. Each ROT is constructed by parsing the SQL into relational algebra expressions that capture the sequence and dependency of logical operations \citep{cyganiak2005relational}. In practice, this conversion is performed using Apache Calcite \citep{begoli2018apache}, which yields a canonical ROT structure referred to as a RelNode. \footnote{\url{https://github.com/apache/calcite}.} The RelNode abstraction refines the logical plan via operator reordering and redundant clause elimination, making it robust to syntactic variations while preserving execution semantics.

Based on RelNode representations, a partial matching strategy is employed to measure the similarity between the relational operator trees of the reference SQL \( R(q^\star) \) and the generated SQL \( R(\hat{q}) \). Let \( \mathcal{N}_{q^\star} \) and \( \mathcal{N}_{\hat{q}} \) denote the sets of nodes in the RelNodes of the reference and generated queries, respectively. A node pair \( (n, n') \) is considered a match, denoted as \( \text{match}(n, n') \), if the two nodes share the same relational operator type and their associated attributes are semantically equivalent.

To determine the best match for a node \( n' \in \mathcal{N}_{\hat{q}} \), we score it against all candidate nodes from \( \mathcal{N}_{q^\star} \), selecting the highest-scoring candidate. Each candidate score \( m^j \) is computed recursively as a weighted sum of node-level similarity and the average similarity of their child nodes:

\vskip-4mm 
$$
m^j = \alpha \cdot m_{\text{self}}^j + (1 - \alpha) \cdot \frac{1}{N} \sum_{i=1}^{N} m_{\text{child}^{(i)}}^j
$$
where $\alpha \in (0, 1)$, and a smaller value of $\alpha$ emphasizes structural alignment by favoring subtree similarity, while a larger value prioritizes node-level matching accuracy.

Based on the above definition, the precision and recall of the matching results are computed as:

\vskip-4mm 
$$
\begin{aligned}
\text{Precision} & = \frac{|\{n \in \mathcal{N}_{\hat{q}} \mid \exists n' \in \mathcal{N}_{q^\star},\ \text{match}(n, n')\}|}{|\mathcal{N}_{\hat{q}}|}
\\
\text{Recall} & = \frac{|\{n \in \mathcal{N}_{q^\star} \mid \exists n' \in \mathcal{N}_{\hat{q}},\ \text{match}(n, n')\}|}{|\mathcal{N}_{q^\star}|}
\end{aligned}
$$

Here, $\text{match}(n, n')$ denotes a binary function that returns 1 if nodes $n$ and $n'$ are matched, and 0 otherwise. In the context of SQL generation, capturing the semantics of the reference SQL is prioritized. The final reward $r_{\text{RelPM}}$ is calculated using a relatively large $\beta$ to emphasize recall as follows:

\vskip-4mm 
\begin{equation}
r_{\text{RelPM}} = \frac{(1 + \beta^2) \cdot \text{Precision} \cdot \text{Recall}}{\beta^2 \cdot \text{Precision} + \text{Recall}}
\end{equation}

\section{FuncEvalGMN}\label{appendix:funcevalgmn_detail}

RelNode presentations are firstly convert into graphs. Nodes in the graphs represent relational operators or expressions, while edges capture both logical execution dependencies and data flow relations. Nodes represent relational operators or expressions, and edges capture execution dependencies and data flows. A GMN \citep{li2019graph} encodes these graphs through three stages: inner-graph message passing, cross-graph semantic alignment, and gated aggregation.

In the inner-graph message passing stage, node embeddings are iteratively updated by aggregating information from their local neighborhoods: 

\vskip-4mm 
\begin{equation} 
\resizebox{0.9\linewidth}{!}{$ 
m_v^{(t+1)} = \sum{u \in N(v)} f_\mathrm{inner}(h_v^{(t)}, h_u^{(t)}, e_{uv}),
$} 
\end{equation} 
where $h_v^{(t)}$ and $h_u^{(t)}$ denote the hidden representations of nodes $v$ and $u$ at step $t$, and $e_{uv}$ is the edge embedding derived from a learned layer.

Next, the cross-graph message passing stage aligns structurally similar components between $\mathcal{G}{\hat{q}}$ and $\mathcal{G}{q^*}$ using cross-attention: 
\vskip-4mm
\begin{align}
\resizebox{0.8\linewidth}{!}{$\displaystyle
\begin{aligned}
r_v^{(t)} &= \mathrm{MLP}(h^{(t)}_v \oplus p_v^{(t)}), \\
\mu_v^{(t+1)} &= \sum_{u \in \mathcal{G}_2(v)} a_{u\to v}(r_v^{(t)} - r_u^{(t)}),
\end{aligned}
$}
\end{align}
where $p_v^{(t)}$ is the positional encoding and $a_{u \to v}$ is the attention weight defined by: 
\vskip-4mm
\begin{equation}
\resizebox{0.8\linewidth}{!}{$\displaystyle
a_{u\to v} = \frac{\exp(s(r^{(t)}_v, r^{(t)}_u))}{\sum_{u' \in \mathcal{G}_2(v)} \exp(s(r^{(t)}_v, r^{(t)}_{u'}))},
$}
\end{equation}

where $s(r_v, r_u) = \frac{r_v \cdot r_u}{\sqrt{d}}$ and $d$ is the dimensionality of node embeddings. This cross-graph attention mechanism enables fine-grained alignment between substructures in different SQL queries. After $T$ propagation steps, the graph-level representation is computed via a gated aggregation mechanism \citep{li2015gated}, which selectively emphasizes salient node features:

\vskip-4mm
\begin{equation}
\resizebox{0.8\linewidth}{!}{$\displaystyle
h_G = \mathrm{MLP}_G\left(\sum_{v \in V} \sigma\left(\mathrm{MLP}_\mathrm{gate}(h_v^{(T)})\right) \odot \mathrm{MLP}(h_v^{(T)})\right)$,
}
\end{equation}

where $\sigma(\cdot)$ is the sigmoid function and $\odot$ denotes element-wise multiplication. To quantify semantic similarity between $\hat{q}$ and $q^\star$, we compute the negative Euclidean distance between their final graph embeddings: 

\vskip-4mm
\begin{equation}
\resizebox{0.6\linewidth}{!}{$
r_{\text{similarity}} = - \left\| h_{G_{\hat{q}}} - h_{G_{q^*}} \right\|_2
$}
\end{equation}

\newpage

\onecolumn

\section{Prompt Strategy of CTE}
We incorporate CTE-style data into the supervised fine-tuning (SFT) phase to encourage the generation of Common Table Expressions (CTEs). Specifically, we augment the training set by rewriting examples from the BIRD-train dataset. Each example is first passed through two prompting stages using GPT-4o: (1) determining whether the original SQL query is complex enough to benefit from CTE rewriting, and (2) if so, performing the actual CTE transformation. We then evaluate the functional equivalence of the rewritten and original SQL queries using Test Suite Accuracy to ensure correctness. This pipeline yields 3,252 rewritten examples (BIRD-train-CTE) and 3,810 unmodified examples that were deemed unnecessary to rewrite. The prompts and an example transformation are shown below.

\subsection{CTE Rewriting Necessity Judgment}
\begin{tcolorbox}[colframe=black!20!white, colback=white, coltitle=black, title=\textbf{Instruction}]
You are provided with the following SQL statement:\\
\{Reference\_SQL\}\\
Evaluate the complexity of the given SQL query. 
\\
Determine if splitting it into Common Table Expressions (CTEs) using WITH x AS clauses would help with understanding.
If the query is complex and CTEs would be beneficial, output \{\{"cte\_necessary": "True"\}\}.
\\
If the query is simple and does not require CTEs (for example, a basic single - table SELECT with simple conditions like SELECT SUM(occurrences) FROM words WHERE LENGTH(word) = 3), output \{\{"cte\_necessary": "False"\}\}.
\end{tcolorbox}

\subsection{CTE Rewriting}
\begin{tcolorbox}[colframe=black!20!white, colback=white, coltitle=black, title=\textbf{Instruction}]
Text-to-SQL is a task of transforming natural language queries into Structured Query Language (SQL) statements. 
\\
Table Schema:\\
\{prompt\_schema\}
\\
Question:\\
\{query\}
\\
You are given the following SQL statement as an response (GroundTruth\_SQL):\\
\{Reference\_SQL\}
\\
Rewrite above SQL query into a Common Table Expression (CTE) format using `WITH x AS` clauses. Break the query into logical steps and use intermediate CTEs for each step.
\\
The response format should be:\\
\{\{"sql": "<CTE-formatted SQL statement>"\}\}
\end{tcolorbox}

\subsection{CTE Rewriting Example}
\begin{table*}[h!]
\centering
\small
\begin{tabular}{lp{12cm}}
\toprule
Type & Content \\
\midrule
Question & What is the shipment ID of the heaviest shipment that Zachery Hicks transported? \\
\midrule
Reference SQL & SELECT T1.ship\_id FROM shipment AS T1 INNER JOIN driver AS T2 ON T1.driver\_id = T2.driver\_id WHERE T2.first\_name = 'Zachery' AND T2.last\_name = 'Hicks' ORDER BY T1.weight DESC LIMIT 1 \\
\midrule
Reference CTE SQL & WITH DriverShipments AS (SELECT T1.ship\_id, T1.weight FROM shipment AS T1 INNER JOIN driver AS T2 ON T1.driver\_id = T2.driver\_id WHERE T2.first\_name = 'Zachery' AND T2.last\_name = 'Hicks'), HeaviestShipment AS (SELECT ship\_id FROM DriverShipments ORDER BY weight DESC LIMIT 1) SELECT ship\_id FROM HeaviestShipment \\
\bottomrule
\end{tabular}
\end{table*}

\clearpage

\section{Prompt Strategy of Text-to-SQL Task}
\label{appendix:prompt strategie}
\begin{tcolorbox}[colframe=black!20!white, colback=white, coltitle=black, title=\textbf{Instruction}]
There are two tables: singer, song.\\
the structure of table singer is as follows:\\
| column name | column type |\\
| ------------ | ------------ |\\
| singer\_id | number |\\
| name | text |\\
| birth\_year | number |\\
| net\_worth\_millions | number |\\
| citizenship | text |\\
singer\_id is the primary key.\\
\\
the structure of table song is as follows:\\
| column name | column type |\\
| ------------ | ------------ |\\
| song\_id | number |\\
| title | text |\\
| singer\_id | number |\\
| sales | number |\\
| highest\_position | number |\\
song\_id is the primary key.\\
The singer\_id is the foreign key, reference to singer\_id in table singer.\\
\\
Question: What is the sname of every sing that does not have any song?\\
The corresponding SQL code is: 
\end{tcolorbox}

\begin{tcolorbox}[colframe=black!20!white, colback=white, coltitle=black, title=\textbf{Reference SQL (Ground Truth SQL)}]
SELECT Name FROM singer WHERE Singer\_ID NOT IN (SELECT Singer\_ID FROM song)
\end{tcolorbox}

\begin{tcolorbox}[colframe=black!20!white, colback=white, coltitle=black, title=\textbf{Example Generated SQL}]
SELECT name FROM singer WHERE NOT singer\_id IN (SELECT singer\_id FROM song)
\end{tcolorbox}

\clearpage

\section{Reasoning Prompt Strategy of Text-to-SQL Task (for Qwen2.5-Coder-7B)}
\label{appendix:reasoning_grpo_prompt}

\begin{tcolorbox}[colframe=black!20!white, colback=white, coltitle=black, title=\textbf{Instruction with Reasoning}]
You are an experienced database expert. Given the database schema and the user's question, your task is to write a single SQL that correctly answers the question. \\
\\
Database schema:\\
There are two tables: manufacturers, products.\\
the structure of table manufacturers is as follows:\\
| column name  |  column type |\\
| ------------ | ------------ |\\
| code | number |\\
| name | text |\\
......
\\
the structure of table products is as follows:\\
......
\\
Question: What is the average price across all products?\\

Respond in the following format:\\
<think>\\
Your detailed and step-by-step thinking path toward finding the correct SQL query \\
</think>\\
<answer>\\
```sql\\
Your predicted SQL query\\
```\\
</answer>\\
Now is your turn to respond in the above format.\\
\end{tcolorbox}

\begin{tcolorbox}[colframe=black!20!white, colback=white, coltitle=black, title=\textbf{Example Generated Response}]
<think>\\
1. **Understand the Task**: The task is to find the total hours of all projects in the database.\\
2. **Identify Relevant Table**: From the schema, the table that contains the project hours is the `projects` table.\\
3. **Identify Relevant Column**: The column in the `projects` table that contains the hours is `hours`.\\
4. **Formulate the SQL Query**: To get the total hours, we need to sum up all the values in the `hours` column of the `projects` table.\\
</think>\\
<answer>\\
```sql\\
SELECT headquarters\\
FROM companies\\
GROUP BY headquarters\\
HAVING COUNT(*) > 2;\\
```\\
</answer>\\
\end{tcolorbox}

\clearpage

\newpage
\section{Case Study: Outcome Reward Comparison}
\label{appendix:case_study}
To illustrate the behavioral differences of reward models under the outcome-based reward setting, we present four case studies comparing responses generated by models trained with SFT, BTRM, EX and GMN rewards. 

\definecolor{LightRed}{RGB}{255,192,192}
\definecolor{LightGreen}{rgb}{0.56, 0.93, 0.56}

\begin{table*}[h!]
    \caption{Sample query, responses from the model trained with Reward model based GMN, BTRM and Execution. If it is an obvious error, we color the text in the error location red.}
    \centering
    \small
    \begin{tabular}{lp{10cm}ll}
        \toprule
        Type & Content & TS \\
        \midrule
Prompt & There are four tables: city, sqlite\_sequence,  country, countrylanguage. & N/A \\
&the structure of table city is as follows:\\
&...\\
&the structure of table sqlite\_sequence is as follows:\\
&...\\
\\
&the structure of table country is as follows:\\
&| column name | column type |\\
&| ------------ | ------------ |\\
&| code | text |\\
&| name | text |\\
&| continent | text |\\
&| region | text |\\
&| surfacearea | number |\\
&| indepyear | number |\\
&| population | number |\\
&| lifeexpectancy | number |\\
&| gnp | number |\\
&| gnpold | number |\\
&| localname | text |\\
&| governmentform | text |\\
&| headofstate | text |\\
&| capital | number |\\
&| code2 | text |\\
&code is the primary key.\\
&\\
&the structure of table countrylanguage is as follows:\\
&...\\

&\\
&Question: Find the government form name and total population for each government form whose \sethlcolor{LightGreen}\hl{average} life expectancy is longer than 72.\\
\midrule
Groundtruth & SELECT sum(Population) , GovernmentForm FROM country GROUP BY GovernmentForm HAVING \sethlcolor{LightGreen}\hl{avg(LifeExpectancy)} > 72 & N/A \\
\midrule
SFT & select governmentform, sum(population) from country where \sethlcolor{LightRed}\hl{lifeexpectancy} > 72 group by governmentform	
& 0 \\
\midrule
BTRM & SELECT governmentform, SUM(population) FROM country WHERE \sethlcolor{LightRed}\hl{lifeexpectancy} > 72 GROUP BY governmentform	& 0  \\
\midrule
EX & SELECT governmentform, SUM(population) FROM country WHERE \sethlcolor{LightRed}\hl{lifeexpectancy} > 72 GROUP BY governmentform & 0 \\
\midrule
GMN &  SELECT governmentform, SUM(population) FROM country GROUP BY governmentform HAVING \sethlcolor{LightGreen}\hl{AVG(lifeexpectancy)} > 72 & 1 \\
\bottomrule
\end{tabular}
\end{table*}

\begin{table*}[h!]
    \caption{Sample query, responses from the model trained with Reward model based GMN, BTRM and Execution. If it is an obvious error, we color the text in the error location red.}    
    \centering
    \small
    \begin{tabular}{lp{10cm}ll}
        \toprule
        Type & Content & TS \\
        \midrule
Prompt & There are six tables: continents, countries, car\_makers, model\_list, car\_names, cars\_data. & N/A \\
& the structure of table continents is as follows: \\
&...\\
& \\
& the structure of table countries is as follows: \\
&...\\
& \\
& the structure of table car\_makers is as follows: \\
&...\\
& \\
& the structure of table model\_list is as follows: \\
&...\\
& \\
& the structure of table car\_names is as follows: \\
& | column name | column type | \\
& | ------------ | ------------ | \\
& | makeid | number | \\
& | model | text | \\
& | make | text | \\
& makeid is the primary key. \\
& The model is the foreign key, reference to model in table model\_list. \\
& \\
& the structure of table cars\_data is as follows: \\
& | column name | column type | \\
& | ------------ | ------------ | \\
& | id | number | \\
& | mpg | text | \\
& | cylinders | number | \\
& | edispl | number | \\
& | horsepower | text | \\
& | weight | number | \\
& | accelerate | number | \\
& | year | number | \\
& id is the primary key. \\
& The id is the foreign key, reference to makeid in table car\_names. \\
& \\
& Question: What is the accelerate of the car \sethlcolor{LightGreen}\hl{make} \sethlcolor{LightGreen}\hl{amc hornet sportabout (sw)}? \\
\midrule
Groundtruth & SELECT T1.Accelerate FROM CARS\_DATA AS T1 JOIN CAR\_NAMES AS T2 ON T1.Id = T2.MakeId WHERE \sethlcolor{LightGreen}\hl{T2.Make} = \sethlcolor{LightGreen}\hl{'amc hornet sportabout (sw)'}; & N/A \\
\midrule
SFT & select t2.accelerate from car\_makers as t1 join cars\_data as t2 on t1.id = t2.id where \sethlcolor{LightGreen}\hl{t1.make} = "amc" and t1.model = \sethlcolor{LightRed}\hl{"hornet sportabout"} & 0 \\
\midrule
BTRM & SELECT cars\_data.accelerate FROM cars\_data INNER JOIN car\_names ON cars\_data.makeid = car\_names.makeid WHERE \sethlcolor{LightRed}\hl{car\_names.model} = \sethlcolor{LightGreen}\hl{'amc hornet sportabout (sw)'} & 0  \\
\midrule
EX & SELECT accelerate FROM cars\_data WHERE id IN (SELECT makeid FROM car\_names WHERE \sethlcolor{LightRed}\hl{model} = \sethlcolor{LightRed}\hl{'sportabout (sw)'}) & 0 \\
\midrule
GMN & SELECT cars\_data.accelerate FROM cars\_data INNER JOIN car\_names ON cars\_data.id = car\_names.makeid WHERE \sethlcolor{LightGreen}\hl{car\_names.make} = \sethlcolor{LightGreen}\hl{'amc hornet sportabout (sw)'} & 1 \\
\bottomrule
\end{tabular}
\end{table*}

\begin{table*}[h!]
    \caption{Sample query, responses from the model trained with Reward model based GMN, BTRM and Execution. If it is an obvious error, we color the text in the error location red. 
    \\ GT Explanation: from the CAR\_NAMES table and the CARS\_DATA table by associating the MakeId with the Id, sort the rows by mpg descending order and take the first record.
    }
    \centering
    \small
    \begin{tabular}{lp{10cm}ll}
        \toprule
        Type & Content & TS \\
        \midrule
Prompt & There are six tables: continents, countries, car\_makers, model\_list, car\_names, cars\_data. & N/A \\
&...\\
& the structure of table model\_list is as follows: \\
& | column name | column type | \\
& | ------------ | ------------ | \\
& | modelid | number | \\
& | maker | number | \\
& | model | text | \\
& modelid is the primary key. \\
& The maker is the foreign key, reference to id in table car\_makers. \\

& \\
& the structure of table car\_names is as follows: \\
& | column name | column type | \\
& | ------------ | ------------ | \\
& | makeid | number | \\
& | model | text | \\
& | make | text | \\
& makeid is the primary key. \\
& The model is the foreign key, reference to model in table model\_list. \\
& \\
& the structure of table cars\_data is as follows: \\
& | column name | column type | \\
& | ------------ | ------------ | \\
& | id | number | \\
& | mpg | text | \\
& | cylinders | number | \\
& | edispl | number | \\
& | horsepower | text | \\
& | weight | number | \\
& | accelerate | number | \\
& | year | number | \\
& id is the primary key. \\
& The id is the foreign key, reference to makeid in table car\_names. \\
& \\
& Question: Which model saves the most gasoline? That is to say, have the maximum \sethlcolor{LightGreen}\hl{miles per gallon}. \\
\midrule
Groundtruth & SELECT T1.Model FROM CAR\_NAMES AS T1 JOIN CARS\_DATA AS T2 ON T1.MakeId = T2.Id ORDER BY T2.mpg DESC LIMIT 1; & N/A \\
\midrule
SFT & select t2.model from cars\_data as t1 join model\_list as t2 on t1.id = t2.modelid group by t1.id order by \sethlcolor{LightRed}\hl{t1.edispl} desc limit 1 & 0  \\
\midrule
BTRM & SELECT model FROM cars\_data ORDER BY mpg DESC LIMIT 1 & 0  \\
\midrule
EX & SELECT model FROM model\_list ORDER BY mpg DESC LIMIT 1 & 0  \\
\midrule
GMN & SELECT car\_names.model \sethlcolor{LightGreen}\hl{FROM cars\_data INNER JOIN car\_names ON cars\_data.id = car\_names.makeid} ORDER BY \sethlcolor{LightGreen}\hl{cars\_data.mpg} DESC LIMIT 1 & 1 \\
\bottomrule
\end{tabular}
\end{table*}

\begin{table*}[h!]
    \caption{Sample query, responses from the model trained with Reward model based GMN, BTRM and Execution. If it is an obvious error, we color the text in the error location red. 
    \\ Error Explanation of SFT: The id equivalents of tv\_channel and cartoon have different meanings and should not be connected simply by using the id.
     \\ Error Explanation of BTRM: Question focused on cartoon is not suitable to use Inner join channel table, no foreign key connection, should not assume that all channels in the cartoon table in the database have a corresponding id stored in tv\_channel.}
    \centering
    \small
    \begin{tabular}{lp{10cm}ll}
        \toprule
        Type & Content & TS \\
        \midrule
Prompt & There are three tables: tv\_channel, tv\_series, cartoon. & N/A \\
& the structure of table tv\_channel is as follows: \\
& | column name | column type | \\
& | ------------ | ------------ | \\
& | id | text | \\
& | series\_name | text | \\
& | country | text | \\
& | language | text | \\
& | content | text | \\
& | pixel\_aspect\_ratio\_par | text | \\
& | hight\_definition\_tv | text | \\
& | pay\_per\_view\_ppv | text | \\
& | package\_option | text | \\
& id is the primary key. \\
& \\
& the structure of table tv\_series is as follows: \\
& | column name | column type | \\
&...\\
& \\
& the structure of table cartoon is as follows: \\
& | column name | column type | \\
& | ------------ | ------------ | \\
& | id | number | \\
& | title | text | \\
& | directed\_by | text | \\
& | written\_by | text | \\
& | original\_air\_date | text | \\
& | production\_code | number | \\
& | channel | text | \\
& id is the primary key. \\
& \\
& Question: find the package option of the tv channel that do not have any cartoon directed by Ben Jones. \\
\midrule
Groundtruth & SELECT package\_option FROM TV\_Channel WHERE id NOT IN (SELECT channel FROM cartoon WHERE directed\_by = 'Ben Jones') & N/A \\
\midrule
SFT & select distinct package\_option from tv\_channel except select t1.package\_option from tv\_channel as t1 \sethlcolor{LightRed}\hl{join cartoon as t2 on t1.id = t2.id} where t2.directed\_by = "Ben Jones" & 0 \\
\midrule
BTRM & SELECT tv\_channel.package\_option FROM tv\_channel EXCEPT SELECT tv\_channel.package\_option FROM tv\_channel INNER \sethlcolor{LightRed}\hl{JOIN cartoon} ON tv\_channel.id = cartoon.channel WHERE cartoon.directed\_by = "Ben Jones" & 0 \\
\midrule
EX & SELECT package\_option FROM tv\_channel WHERE NOT id IN (SELECT \sethlcolor{LightRed}\hl{directed\_by} FROM cartoon WHERE directed\_by = 'Ben Jones') & 0 \\
\midrule
GMN & SELECT package\_option FROM tv\_channel WHERE NOT id IN (SELECT channel FROM cartoon WHERE directed\_by = 'Ben Jones') & 1 \\
\bottomrule
\end{tabular}
\end{table*}

\end{document}